\documentclass[10pt]{article}
\usepackage[accepted]{rlj} %

\usepackage{amssymb}            %
\usepackage{mathtools}          %
\usepackage{mathrsfs}           %
\usepackage{graphicx}           %
\usepackage{subcaption}         %
\usepackage[space]{grffile}     %
\usepackage{url}                %
\usepackage{lipsum}             %

\usepackage{color, float}
\usepackage{xspace}
\usepackage{booktabs} %

\usepackage[capitalize,noabbrev,nameinlink]{cleveref}
\usepackage{listings}
\renewcommand{\cite}{\citep}

\newcommand{\D}{\mathcal{D}}

\newcommand{\N}{\mathcal{N}}

\renewcommand{\H}{\mathcal{H}}

\newcommand{\rint}{r^\mathrm{int}}
\newcommand{\oni}{\texttt{ONI}\xspace}
\newcommand{\oniretrieval}{{\small \texttt{ONI-retrieval}}\xspace}
\newcommand{\onicls}{{\small\texttt{ONI-classification}}\xspace}
\newcommand{\oniranking}{{\small\texttt{ONI-ranking}}\xspace}
\newcommand{\ellm}{{\small\texttt{ELLM}}\xspace}
\newcommand{\ellmbow}{{\small\texttt{ELLM-BoW}}\xspace}

\definecolor{codegreen}{rgb}{0,0.6,0}
\definecolor{codegray}{rgb}{0.5,0.5,0.5}
\definecolor{codepurple}{rgb}{0.58,0,0.82}
\definecolor{backcolour}{rgb}{0.95,0.95,0.92}

\lstdefinestyle{mystyle}{
    backgroundcolor=\color{backcolour},   
    commentstyle=\color{codegreen},
    keywordstyle=\color{magenta},
    numberstyle=\tiny\color{codegray},
    stringstyle=\color{codepurple},
    basicstyle=\ttfamily\footnotesize,
    breakatwhitespace=false,         
    breaklines=true,                 
    captionpos=b,                    
    keepspaces=true,                              
    showspaces=false,                
    showstringspaces=false,
    showtabs=false,                  
    tabsize=2
}

\title{Online Intrinsic Rewards for Decision Making Agents from Large Language Model Feedback}
\setrunningtitle{Online Intrinsic Rewards for Decision Making Agents from Large Language Model Feedback}

\author{Qinqing Zheng\textsuperscript{$1\dagger$}, 
Mikael Henaff\textsuperscript{$1\dagger$}, 
Amy Zhang\textsuperscript{$1,2$},
Aditya Grover\textsuperscript{$3$}, \\
Brandon Amos\textsuperscript{$1$}
}

\affiliations{
$^{1}$\textbf{Meta} \;
$^{2}$\textbf{UT Austin} \;
$^{3}$\textbf{UCLA}
\\
$^\dagger$ Equal contribution}

\contribution{
This paper presents a method, system design and open-source codebase for learning intrinsic rewards from LLM feedback in an online manner, which scales to high-throughput settings.
}
{
Several methods for producing intrinsic rewards from LLM feedback have been proposed in prior work. However, they either do not scale to high-throughput settings, or they require a large and diverse offline dataset. Our method works in high-throughput settings without requiring offline data.
}


\summary{
Automatically synthesizing dense rewards from natural language descriptions is a promising paradigm in reinforcement learning (RL), with applications to sparse reward problems, open-ended exploration, and hierarchical skill design. Recent works have made promising steps by exploiting the prior knowledge of large language models (LLMs). 
However, these approaches suffer from important limitations: they are either not scalable to problems requiring billions of environment samples, due to requiring LLM annotations for each observation, or they require a diverse offline dataset, which may not exist or be impossible to collect. 
In this work, we address these limitations through a combination of algorithmic and systems-level contributions. We propose \oni, a distributed architecture that simultaneously learns an RL policy and an intrinsic reward function using LLM feedback. Our approach annotates the agent's collected experience via an asynchronous LLM server,  which is then distilled into an intrinsic reward model. We explore a range of algorithmic choices for reward modeling with varying complexity, including hashing, classification, and ranking models. 
Our approach achieves state-of-the-art performance across a range of challenging, sparse reward tasks from the NetHack Learning Environment in a simple unified process, solely using the agent's gathered experience, without requiring external datasets.
}

\begin{document}

\maketitle  %

\begin{abstract}
Automatically synthesizing dense rewards from natural language descriptions is a promising paradigm in reinforcement learning (RL), with applications to sparse reward problems, open-ended exploration, and hierarchical skill design. Recent works have made promising steps by exploiting the prior knowledge of large language models (LLMs). 
However, these approaches suffer from important limitations: they are either not scalable to problems requiring billions of environment samples, due to requiring LLM annotations for each observation, or they require a diverse offline dataset, which may not exist or be impossible to collect. 
In this work, we address these limitations through a combination of algorithmic and systems-level contributions. We propose \oni, a distributed architecture that simultaneously learns an RL policy and an intrinsic reward function using LLM feedback. Our approach annotates the agent's collected experience via an asynchronous LLM server,  which is then distilled into an intrinsic reward model. We explore a range of algorithmic choices for reward modeling with varying complexity, including hashing, classification, and ranking models. 
Our approach achieves state-of-the-art performance across a range of challenging tasks from the NetHack Learning Environment, while removing the need for large offline datasets required by prior work.
We make our code available at %
 \url{https://github.com/facebookresearch/oni}. \looseness=-1
\end{abstract}

\section{Introduction}
\label{sec:intro}
Reward functions are central to reinforcement learning (RL), and are often assumed to be given as part of the problem definition \citep{Sutton1998}. 
These functions are written to describe the task at hand, and often involve tradeoffs between ease of task definition and ease of policy optimization.  For example, assigning a reward of $+1$ for solving the task and $0$ otherwise is simple to define and accurately reflects the task goal, but is difficult to optimize due to providing zero gradients almost everywhere. 

These difficulties have motivated the use of intrinsic rewards to aid policy optimization
\citep{randlov1998shaping, ng1999rewardshaping, sorg2010internal, singh2010intrinsically}.  The reward designer can include additional reward shaping terms to create a denser learning signal, which can reflect task progress or guide the agent towards intermediate goals. However, designing intrinsic rewards can be remarkably challenging~\citep{booth23perils,  ibrahim2024comprehensive} 
and places increased demands on human experts to provide task-specific knowledge. \looseness=-1

\begin{figure}[t]
  \centering
  \includegraphics[width=\textwidth]{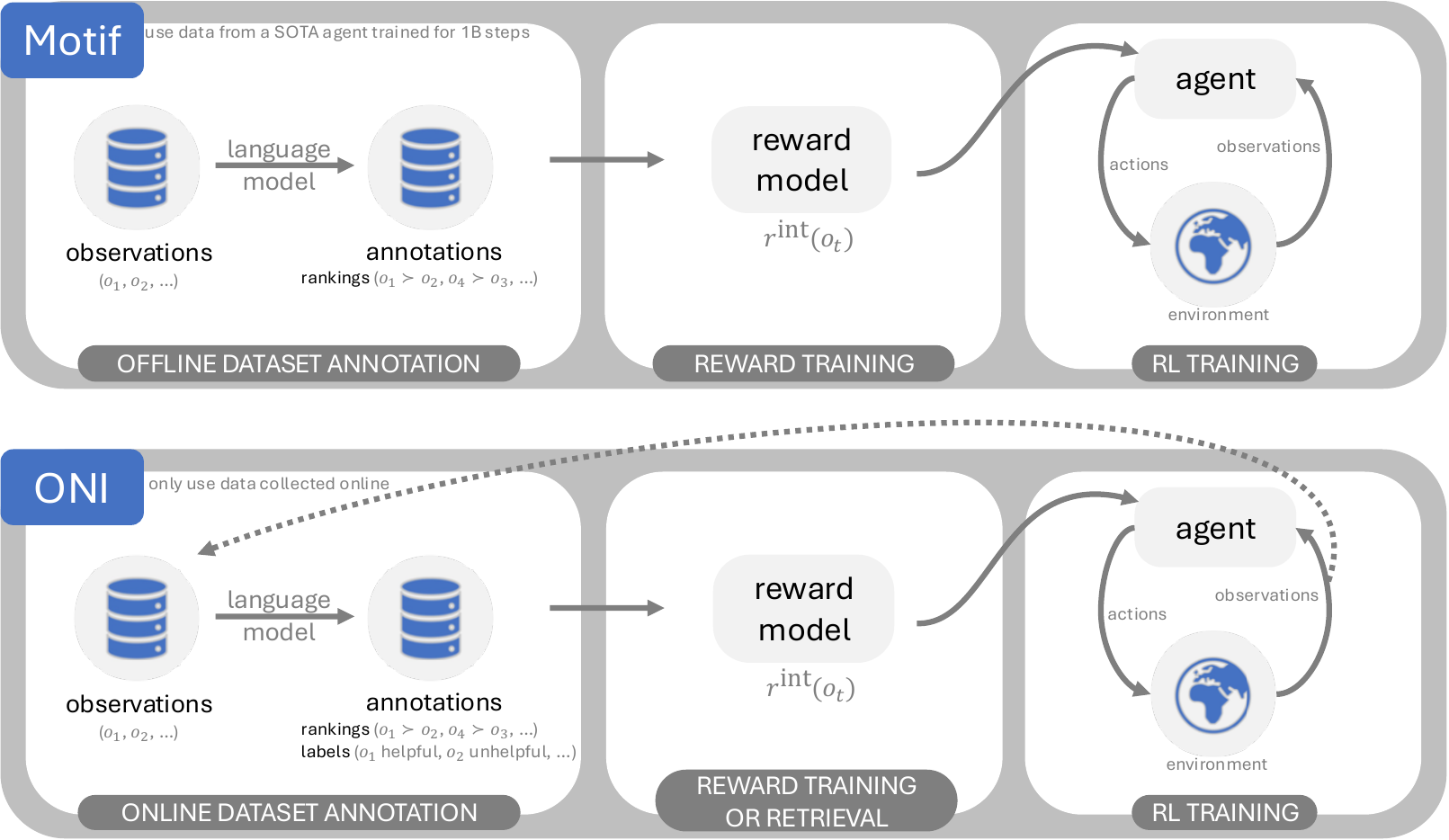}
  \caption{Overview of Motif \citep{klissarovdoro2023motif} and ONI.}
  \label{fig:overview}
\end{figure}

Recently, several works have been proposed to leverage the vast prior knowledge encoded in large language models (LLMs) to automate the reward design process, based on a task description in natural language.
They can be broadly categorized into two families:
\begin{enumerate}[itemsep=0pt, topsep=0pt, leftmargin=0pt]
\vspace*{-3pt}
\item[] \textbf{1. Generating the reward function's code by LLM.}
  A number of methods have been proposed to automatically generate
  executable code that computes the reward directly~\citep{ma2023eureka, xie2023text2reward, yu2023language, li2024auto}. 
  While they have demonstrated success in complex continuous control tasks, they either require access to environment source code to include in the prompt, or a detailed description of input parameters and reward function templates.
  Furthermore, they are limited to reward functions compactly expressible via code,  describing explicit logic, and it is unclear how these approaches can easily process high-dimensional state representations such as images, or semantic features such as natural language. 
\item[] \textbf{2. Generating reward values by LLMs.} 
  Motif~\citep{klissarovdoro2023motif, klissarov2024maestromotif} is a typical example of this category.
  It ranks the captions of pairs of observations using an LLM and distills these preferences into a parametric reward model. Motif does not require access to environment source code nor numerical state representation, can process semantic input features, and can scale to problems requiring billions of environment samples.
  Nevertheless, it also suffers from two important limitations. First, it requires a diverse, pre-existing dataset of captioned observations which are used to elicit preferences from the LLM. In many situations, such a dataset might not exist, and collecting it can increase the sample complexity.
  More importantly, collecting a diverse dataset often requires a non-trivial reward function that is feasible to optimize,
  which is the primary problem we aim to solve with intrinsic reward functions in the first place. 
  Second, it involves a complex three-stage process, which sequentially annotates observations using an LLM, trains a reward model, and finally trains an RL agent. This is still time-consuming, given that the LLM annotation process can take several days' worth of GPU hours,
  and is done prior to training the reward model and RL agent.
  Alternatively, \citet{chu2023accelerating} query the LLM to directly label observations as having high or low reward at each timestep. 
  However, querying an LLM for every observation is computationally infeasible for many RL applications, which involve 
  millions or billions of observations.  \looseness=-1
\end{enumerate}
As a consequence, it would be desirable to have \emph{an integrated solution} that offers: 
\begin{enumerate}[leftmargin=15pt, itemsep=0pt]
    \item[(1)] \emph{concurrent and fast online learning of both the intrinsic rewards and the policy 
that requires no external data nor auxiliary reward functions,}
    \item[(2)] \emph{expressible reward functions that can capture semantic features
that are difficult to process with compact executable code.}
\end{enumerate}
In this work, we present \oni, a distributed online intrinsic reward and agent learning system --- \Cref{fig:overview} overviews the high-level components.
\oni removes the dependency on external datasets beyond the agent's own experience and enables large-scale RL training with ease. 
\oni assumes access to captions of observations, similar to previous work~\cite{klissarovdoro2023motif, chu2023accelerating}. 
The captions of collected observations are annotated online by an asynchronous LLM server, and both the policy and intrinsic reward model are simultaneously updated using the LLM's feedback.
Such a learning framework allows us to easily instantiate different algorithmic choices for synthesizing LLM feedback. Specifically, we explore three methods: the first one is retrieval-based and simply hashes the annotations; 
the second builds a binary classification model to distill the sentiment labels returned by the LLM;
and the third sends pairs of captions to the LLM server for preference labeling and learns a ranking model, similar to Motif. \looseness=-1
We demonstrate that \oni is able to match Motif's performance across a range of challenging, sparse rewards from the NetHack Learning Environment (NLE)~\citep{kuettler2020nethack}, solely using the agent's gathered experience in a single, unified process. We also open-source our code to facilitate future work on LLM-based intrinsic motivation in settings where large offline datasets are not available. 

\section{Background}

We consider a partially observed Markov decision process (POMDP) setting where the problem is defined by $\mathcal{M} = (\mathcal{S}, \mathcal{A}, \mathcal{O}, p_0, P, O, r, \gamma)$. 
At each episode, an initial state $s_0 \in \mathcal{S}$ is sampled from the initial state distribution $p_0$. 
At each time step $t$, the agent observes $o_t \in \mathcal{O}$
which is computed by the emission function $O(s_t)$,  and takes an action $a_t \in \mathcal{A}$. 
This action causes the environment to transition to a new state, $s_{t+1} \sim p(s_t, a_t)$. A new observation $o_{t+1}$ and reward $r_{t+1} = r(o_{t+1})$ are given to the agent, and the process continues. The goal of the agent is to learn a policy $\pi: \mathcal{O} \rightarrow \Delta(\mathcal{A})$ which maximizes the expected return $\mathbb{E}_{\pi}\left[\sum_t \gamma^t r_t\right]$. In this work, we additionally assume each observation $o_t$ includes a textual caption $c_t$, which could be empty. 
For observation spaces without textual captions, $c_t$ could in principle be provided by a captioning model as well. \looseness=-1

In many situations, the extrinsic environment reward $r$ is sparse, and the resulting return objective is challenging to optimize. We therefore consider methods which make use of an auxiliary \textit{intrinsic} reward $\rint$ and define a composite surrogate reward:
\begin{equation}
\bar{r}(o_t) =  r(o_t) + \beta \cdot \rint(o_t).
\label{eq:composed_reward}
\end{equation}
A key research question is how to define or learn the intrinsic reward $\rint$. 
A first option is to manually define $\rint$ based on task-specific goals, for example a measure of the distance between the agent's current state and the goal state. However, handcrafting the intrinsic reward function can require significant domain knowledge and must be redone for each new task.
A second option is to define $\rint$ to measure some notion of observation novelty, which encourages the agent to systematically explore the environment. This can work well in smaller environments, but fails in ones that cannot be exhaustively explored in a tractable amount of time. 
A third class of methods, which we focus on in this work, leverage LLMs to automatically synthesize $\rint$ to reflect prior knowledge about the task. We discuss all three classes of methods in more detail in \Cref{sec:related}.

\section{Online Intrinsic Rewards}

\subsection{System design: distributed PPO with LLM annotations}
\label{sec:system}
This section outlines our system for learning online intrinsic rewards
alongside the policy optimization.
The engineering and design are an important piece of our research, as they enable scaling to the high-throughput settings we are interested in.
The rest of our experimental studies are conducted within this system and influenced by the throughput of its interacting components.

\begin{figure*}[t]
  \centering
  \includegraphics[width=0.9\textwidth]{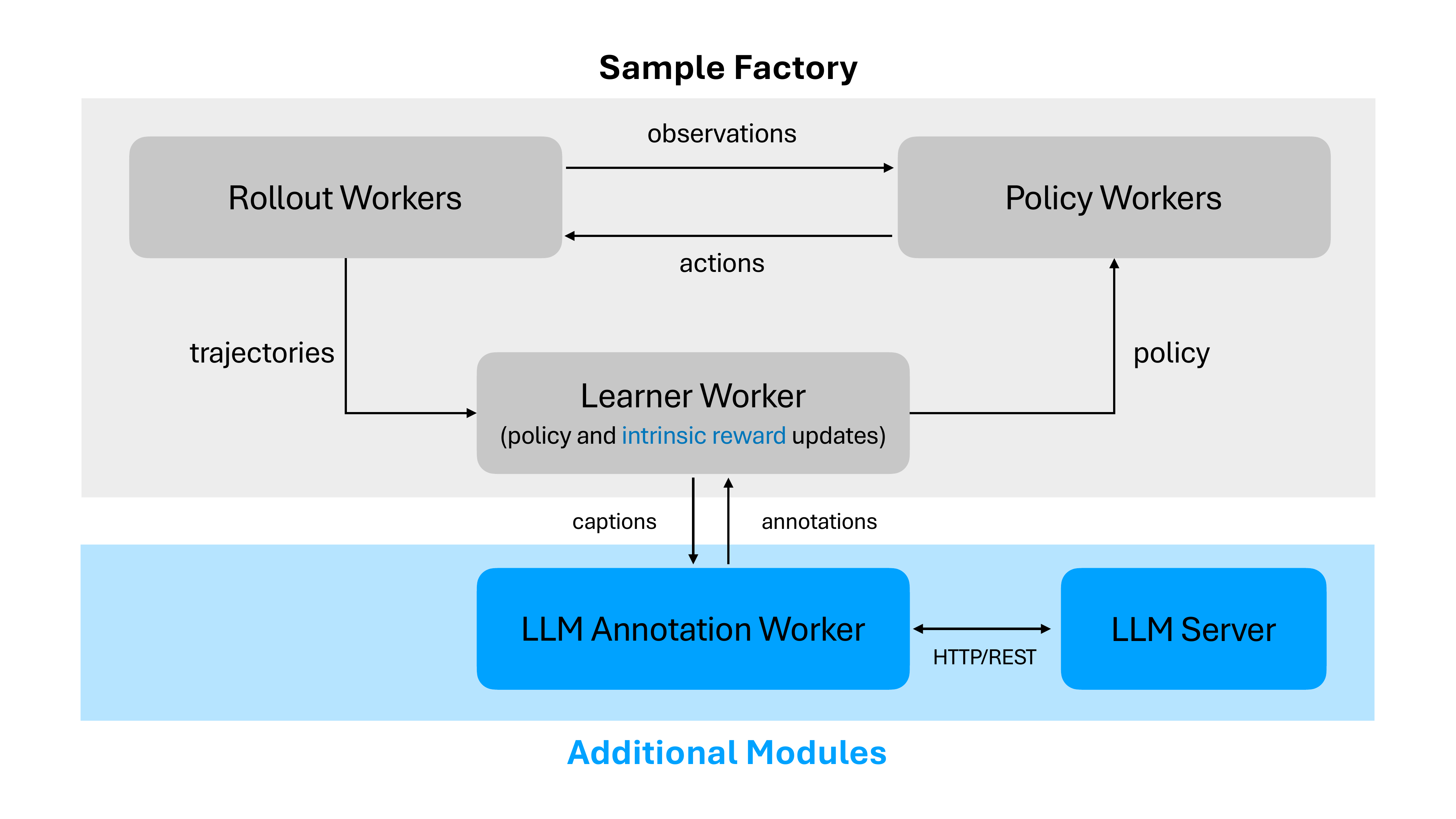}
    \vspace{-5mm}
  \caption{Overall system diagram of \oni. Our additions to Sample Factory are highlighted in blue.
    We added an asynchronously executing LLM server and learned
    reward function, and connect them back into the
    main learning process in a way that does not
    hurt the overall throughput of the policy and value learning.
  }
  \label{fig:system}
  \vspace{-5mm}
\end{figure*}

Our core system, illustrated in \cref{fig:system},
is built on top of the Sample Factory library v1.0~\citep{petrenko2020sf}
and their asynchronous variant of proximal policy optimization
\citep{schulman2017proximal}, referred to as APPO.
APPO operates on a single machine and concurrently runs many
environment instances while asynchronously updating policy and value
estimates with the usual PPO rules, and
adequately handles policy staleness and data transfers between these components.
Concretely for NetHack, by running 480 environment instances
APPO collects approximately 32k
environment interactions per second on a Tesla A100-80GB GPU with 48 CPUs.
For online reward learning via LLM annotations in this system\footnote{
The prior work by \citet{klissarovdoro2023motif} in the offline setting
was able to connect their intrinsic reward function into Sample Factory's
APPO implementation with a few lines of code, as it just needed to load
the  PyTorch module of the learned reward function.},
we added 1) an LLM server hosted on a separate node,
2) an asynchronously running process that passes observation captions to
the LLM server via HTTP request, 
3) a hash table that stores the captions and corresponding LLM annotations, and
4) learning code that dynamically updates a reward model 
that learns on these annotations.
Without being asynchronous, these components 
have the potential to block the concurrent execution and
can reduce the throughput of the overall system,
because calling into an LLM and updating the reward model are time-consuming.
We added them in a way that retains most of the throughput (!), 
approximately 80-95\% of the original:
the average throughput is 30k environment interactions per second if we do not train
any additional reward model to distill the LLM annotations, 
and 26k if a classification-based reward model is learned at the same time (see \Cref{sec:methods} for the description of those approaches).
The throughput of LLM annotation depends on the actual LLM and prompt we are using. 
For the reader's reference, when hosting a  LLaMA-3.1-8B server on a Tesla A100-80GB GPU
using the prompt displayed in \Cref{app:prompts}, \oni annotates
approximately 810k NetHack captions over the whole training process (2B environment steps, 19-20 hours).
The rest of this section overviews the relevant components and
how we have modified them.

\textbf{Background: APPO's Distributed Execution and Shared Memory}
APPO has three main types of workers:
\begin{enumerate}[itemsep=0pt, topsep=0pt]
\item the \textbf{learner worker} that coordinates most of the operations, updates
the policy and value models, and sends the latest policy ID to the
other workers so that they can retrieve it;
\item the \textbf{rollout workers} that run copies of the environment, execute
actions in them and save the observations; and
\item the \textbf{policy workers} that query the policy on new states. 
The policy workers are separate from the rollout workers so they can efficiently batch
across observations.
\end{enumerate}

The workers here are individual processes forked off from a parent process.
These all have shared CPU and GPU memory buffers, and
efficiently communicate mostly via Python's multiprocessing queues,
using pointers to locations in the shared memory.
We next describe our new LLM worker, and how we connect it back into the learner.

\textbf{Our New Asynchronous LLM Worker and Remote LLM Server}
This worker has input and output queues 
that communicate
back with the learner process, which we use in a way that does not block the
main execution of the system.
The LLM worker awaits new observations to label from the input queue, 
formats them into
prompts, calls into an LLM, and returns the annotations in the
output queue. 
Additionally, the prompt templates and other LLM options are configured by the LLM worker.
We use an annotation and message format that support all the intrinsic reward
labels that we consider in \Cref{sec:methods}.
As Sample Factory already utilizes most of the free CPU and GPU
capacity of the system, we opted to call into the LLM via an
HTTP/REST interface rather than loading it in this process directly.
The communication here adds minimal overhead, and also avoids
the need to coordinate the shared memory of multiple GPUs between the main
APPO code and LLM.
Alongside every APPO run, we allocate an unloaded machine
with a new instance of vLLM \citep{kwon2023efficient} to exclusively
serve the queries from the RL run. 
Our system also allows using the same node for both processes if computational resources allow, which further reduces communication costs.
\looseness=-1

One important design decision is that the asynchronous LLM server is only able to
process a small percentage of the overall observations encountered \footnote{often $\leq 0.04\%$
for a LLaMA-3.1-8B instance using a single A100-80GB GPU, or $\leq 0.02\%$ when using a V100-32GB GPU},
because the environment rollout workers have a much higher throughput than the LLM.
An alternative design would be to block the main APPO components and wait
for the LLM to label more messages, but we find it more realistic to continue
running the policy and label messages as the LLM throughput allows.
This also creates the problem of needing to decide on what messages to
send to the LLM: a last-in-first-out queue (LIFO), or some uncertainty-based selection.
For this work, we use LIFO for simplicity, but note that it would be interesting to investigate alternate approaches in future work.
\looseness=-1

\textbf{Our Modified Learner}
Lastly, we connect this new LLM worker back into the main learner,
and dynamically learn a reward model on the annotations obtained from it.
To do this, we modified the two threads of the learner worker:
\begin{enumerate}[itemsep=0pt, topsep=0pt]
\item the \textbf{training thread} is the main thread that
a) aggregates the new trajectories from the latest environment interactions;
b) decides which new observations to send to the LLM;
c) updates the policy, value, and (now) our intrinsic reward model.
\item The \textbf{feedback processing thread} (in fact, the initial thread of the learner worker) receives the latest
  annotations from the LLM worker and updates the dataset (or hash table)
  that the reward model is trained on.
  This thread also initializes the reward model at the beginning. \looseness=-1
\end{enumerate}
To prevent overfitting the limited amount of annotations in the early stage of training,
the training thread only continuously updates the reward model after we receive $25$k annotations. 
Before that, the feedback processing thread runs a few updates of the reward model every time we receive new annotations.
Now that we have a flexible way of annotating messages and fitting a reward
model on top of them, we turn to defining the types of intrinsic reward functions.

\subsection{Intrinsic Reward Functions}
\label{sec:methods}
\oni offers flexible algorithmic choices for querying an LLM and distilling its feedback. In this work, we consider the following
three methods.

\textbf{Retrieval} The simplest approach we consider uses binary labeling and retrieval. 
The LLM is asked to assign a binary label $y_i \in \{0, 1\}$ indicating whether a caption $c_i$ is ``helpful'' or ``unhelpful'' for making progress on the task. The learner worker maintains a hash table $\mathcal{H}$ to store labeled pairs $(c_i, y_i)$, managed by the feedback processing thread.
In the training thread, each time the RL agent receives an observation $o_t$ with caption $c_t$, we check if $c_t \in \mathcal{H}$ and
the intrinsic reward is defined as:
\begin{equation}
    \rint(o_t) = \begin{cases}
    \mathcal{H}(c_t) & \mbox{ if } c_t \in \mathcal{H}\\
    0 & \mbox{ if } c_t \notin \mathcal{H}
    \end{cases}
\end{equation}
If $c_t$ is unlabeled, it is placed into a last-in-first-out (LIFO) queue $Q$ managed by the LLM annotation process,
and then sent to the LLM server. 
The LLM continuously processes elements $c_i$ from $Q$ and returns their labels $y_i$ to the data processing thread of the learner worker, where the pairs $(c_i, y_i)$ are added into $\mathcal{H}$. As described in \Cref{sec:system}, the training thread and the feedback processing thread run asynchronously. Therefore, editing the hash table does not slow down policy training. %
This retrieval-based approach does not generalize to observations with unlabeled captions. However, the resulting intrinsic reward is simple and hyperparameter-free, and may work well when the set of captions belongs to a relatively small set.

\textbf{Classification} The second approach we consider is based on binary labeling together with training a classification model. Similarly to above, we label observation captions $c_i$ with labels $y_i \in \{0, 1\}$ indicating whether they are helpful or unhelpful via LLM. 
We simultaneously train a binary classification model
to predict $y_i$ from $o_i$. More precisely, we model $P(y = 1|o)$ by $\rint_\phi: \mathcal{O} \rightarrow [0, 1]$,
which is then used to compute the binary intrinsic reward by thresholding it at $\eta$: \looseness=-1
\begin{equation}
 r^\mathrm{int}(o_t) = \mathbb{I}[\rint_\phi(o_t) > \eta],
\end{equation}
where $\mathbb{I}$ is the indicator function. We study the impact of $\eta$ in \Cref{sec:ablations}. Unlike the previous approach, this method has the potential to generalize to observations whose captions are similar, but not identical, to the captions labeled by the LLM. However, like the previous approach, it will assign a same reward to observations which are slightly positive (such as finding a few gold pieces in NetHack) and very positive (finding hundreds of gold pieces or a rare artifact). 

\textbf{Ranking} The third approach we consider is based on ranking observations via pairwise classification, which is the approach taken by Motif. Here, pairs of observations $(o_1, o_2)$ are sent to the LLM, which returns a preference label $y \in \{1, 2, \varnothing \}$ indicating whether $o_1$ or $o_2$ is more desirable for accomplishing the task, or if they are equivalent. 
A reward model $\rint_\phi: \mathcal{O} \rightarrow \mathbb{R}$ is trained by minimizing the negative log-likelihood :
\begin{equation}
 \resizebox{0.9\columnwidth}{!}{
$\displaystyle - \mathbb{E}_{(o_1, o_2, y) \sim \H } \Big[ \Big(\mathbb{I}[y = 1] + \tfrac{1}{2} \mathbb{I}[y = \varnothing] \Big) \log P_\phi(o_1 \succ o_2) + 
  \Big(\mathbb{I}[y = 2] + \tfrac{1}{2} \mathbb{I}[y = \varnothing]\Big) \log P_\phi(o_1 \prec o_2) \Big]
$}
\end{equation}
where we use average log-likelihood when $y=\varnothing$\footnote{In the equivalent cross entropy minimization formulation, this amounts to using a uniform target $[\frac12, \frac12]$ when $y=\varnothing$.}
and use the Bradley-Terry model~\cite{bradley1952rank}
$P_\phi (o_1 \succ o_2) = \exp\big( \rint_\phi(o_1)\big) / \big[  \exp\big({\rint_\phi(o_1)}\big) + \exp\big( \rint_\phi(o_2)\big) \big]$.
Motif computes the mean $\mu_\D$, standard deviation $\sigma_\D$, and a fixed quantile $\nu_\D$ of $\rint_\phi$ over the offline dataset of annotations $\D$. During RL training, it normalizes and thresholds $\rint_\phi$ to give the reward 
\begin{equation}
\label{eq:motif_thresholding}
 \rint_\text{motif}(o_t) = \mathbb{I}[(\rint_\phi(o_t) - \mu_\D) / \sigma_\D > \nu_\D] \cdot (\rint_\phi(o_t) - \mu_\D) / \sigma_\D
\end{equation}
For \oni-ranking, applying these steps directly is not possible since the annotation dataset is continuously changing. 
We thus replace $(\mu_\D, \sigma_\D)$ by a running mean and standard deviation $(\mu, \sigma)$ computed over the experience, and replace the quantile $\nu_\D$ by a quantile of the standard normal.

\section{Related Work}
\label{sec:related}
\textbf{Exploration Based on Novelty Bonuses} Learning from sparse or otherwise difficult-to-optimize reward functions is a long-standing problem in reinforcement learning.
There is a large body of work which defines intrinsic rewards based on novelty bonuses \citep{schmidhuber1991possibility, E3, Rmax, stadie2015incentivizing,  PseudoCounts, ICM, RND, max, RIDE, ecoffet2019go, PCPG, NovelD, E3B,lu2024intelligent}. These methods tend to make minimal assumptions about the task at hand, operate online without requiring external data, and sometimes come with theoretical guarantees. However, since they are fundamentally \textit{tabula-rasa}, they must rediscover much of the structure in the task that might already be encoded as prior knowledge in an LLM. Therefore, they tend to have difficulty exploring environments of very high complexity, such as NetHack, in a tractable amount of time \citep{klissarovdoro2023motif}.
\looseness=-1

\textbf{LLM-aided Reward Design}
In addition to Motif~\citep{klissarovdoro2023motif}, several works have sought to leverage the prior knowledge encoded in LLMs to produce intrinsic rewards.
Eureka \citep{ma2023eureka}, Auto-MC \citep{li2024auto}, L2R~\citep{yu2023language} and Text2Reward~\citep{xie2023text2reward}
all use LLMs to generate executable code which computes intrinsic rewards from the underlying environment state, conditioned on a task description. The generated reward function code is then iteratively improved based on aggregate statistics from agents trained with the current reward.
However, a disadvantage with intrinsic rewards represented as code is that they require access to an interpretable underlying state representation, and it is unclear how to leverage non-numerical features such as those provided by unstructured language captions.
The works of \citet{kwon2023reward, chu2023accelerating} also successfully used LLMs conditioned on task descriptions to directly generate binary rewards in an online manner, and did not train a reward model. This was possible due to evaluating on environments and tasks which could be solved with a relatively small number of observations, whereas we consider complex open-ended environments with billions of observations so that labeling them all with an LLM would be computationally infeasible. Also of note is the work of \citet{wu2024read}, which additionally conditioned LLMs on user manuals to define the intrinsic reward. \looseness=-1

\textbf{Goal-conditioned Reward Design}
A different approach to reward function design is to define rewards as the distance between the agent's current state and  the goal. For example, one line of work learns a state embedding in a self-supervised manner which converts geodesic distances in the original space to Euclidean distances in feature space  \citep{wu2018the, DBLP:journals/corr/abs-2107-05545, gomez2024properlaplacianrepresentationlearning}.
Another line of work of \citep{fan2022minedojo, rocamonde2023vision, adeniji2023languagerewardmodulationpretraining,kim2024guide} leverages pretrained image and text encoders, and defines rewards to be some measure of similarity between embeddings of visual observations and embeddings of textual task descriptions.
Using a question generation and answering system, \citet{carta2022eager} extract auxiliary objectives from the goal description and construct intrinsic rewards.
An interesting combination of goal-conditioned and LLM-aided reward design is the \ellm approach introduced in \citet{pmlr-v202-du23f}, which generates candidate goals and uses the distance in LLM embedding space to define the reward. 
This approach shares the limitations of the works of \citet{kwon2023reward, chu2023accelerating} discussed in the previous section, in that it requires an LLM call for each agent observation, which becomes computationally infeasible in high-throughput settings involving billions of observations.
We discuss more works that utilize LLM for RL broadly in \Cref{app:additional_related_work}.

\section{Experiments}
\label{sec:expr}
\textbf{Environment} We use the NetHack Learning Environment (NLE) \citep{kuettler2020nethack} as our experimental testbed, since it is one of the most challenging open-ended, long horizon and sparse reward environments available, and was also used as the main environment in the prior work we compare to. NetHack is a classic dungeon crawling game which presents a number of interesting challenges for RL agents: it is procedurally generated, requiring generalization; rewards are sparse for most tasks, requiring exploration; the environment is partially rather than fully observable; transitions are naturally stochastic; episodes are very long, requiring tens to hundreds of thousands of steps to win the game; the dynamics are highly complex, involving large numbers monsters, objects, non-player characters and other entities. Succeeding in the game requires mastering and balancing diverse behaviors including exploration, resource management, object use, combat, puzzle solving and skill progression. We used the environment settings from \citep{klissarov2024maestromotif}, which include macros for eating, quaffing potions, enhancing skills and casting spells. \looseness=-1

\textbf{Tasks and Metrics} 
We evaluate our agents in two ways: how well they are able to succeed in tasks which have a (typically sparse) extrinsic reward, and how well they are able to progress in the game using the intrinsic reward only. For the former, we use one dense reward task and three sparse reward tasks used in prior work \citep{kuettler2020nethack, klissarovdoro2023motif}, listed below.
\begin{enumerate}[topsep=0pt, itemsep=0pt]
    \item The \texttt{Score} task treats the in-game score\footnote{\url{https://nethackwiki.com/wiki/Score}} as a dense extrinsic reward.
    \item The \texttt{Oracle} task requires finding the in-game Oracle character, which resides deep in the dungeon. The agent receives a reward of 50 if it manages to reach the Oracle, zero otherwise. 
    \item The \texttt{StaircaseLvl3} and \texttt{StaircaseLvl4} tasks require reaching the third or fourth level staircase and zero otherwise---this requires exploring multiple levels in order to find staircases which lead deeper into the dungeon, while fighting or escaping monsters to survive. 
\end{enumerate}
Previous work found that the \texttt{Oracle} task can be solved in unexpected ways via reward hacking; however, this is still a challenging sparse reward problem and we include it for completeness. 
We also train agents with the intrinsic reward only and measure the game progress via four metrics: 

{
\hspace*{15pt} \emph{experience level, \; dungeon level, \; gold, \; and \, scout (number of unique locations explored)}. 
}

Removing the extrinsic reward gives us a clearer picture of what the intrinsic rewards are prioritizing.

\textbf{Methods and Hyperparameters} 
We instantiate \oni with the three reward functions described in \Cref{sec:methods}. We name the three approaches \oniretrieval, \onicls and \oniranking, respectively.
For policy learning, we use the Chaotic Dwarven GPT5 (CDGPT5) architecture \citep{CDGPT5} used in prior work \citep{piterbarg2023nethack, kurenkov2023katakomba, klissarovdoro2023motif}. Architecture details can be found in \Cref{sec:architecture-details}.  
All the methods are trained with two billion $(2 \times 10^9)$ environment steps.  We train \onicls with 
with classification threshold $\eta=0.5$, where the intrinsic reward coefficient is $\beta = 0.1$ for the \texttt{Score} task and $\beta=0.4$ for all the other sparse-reward tasks and the reward-free agent. Similarly, \oniretrieval uses $\beta = 0.1$ for \texttt{Score} and $\beta=0.5$ for the others. 
\oniranking is trained with $\beta=0.05$ and $\nu_\N = 0$. Full details of the training process can be found in \Cref{sec:rl-details}.

\textbf{LLMs} We use the LLaMA-3 herd of models \citep{dubey2024llama3herdmodels} as our LLMs. All prompts are listed in \Cref{app:prompts}. We initially reran the Motif baseline using the official code\footnote{\url{https://github.com/facebookresearch/motif}} and compared the performance of LLaMA-3.1-70B-Instruct and LLaMA-3.1-8B-Instruct (see \Cref{app:llm_model_size}). We did not observe a significant difference in their performance on the \texttt{Oracle} or \texttt{Score} tasks.
Therefore, we use LLaMA-3.1-8B-Instruct in our subsequent experiments to reduce computation. \looseness=-1

\textbf{Baselines} We compare to 3 baselines: i) agents trained with extrinsic reward alone, ii) Motif~\citep{klissarovdoro2023motif},
 and iii) a variant of the \ellm algorithm \cite{du2023guiding}. It is not feasible to directly apply \ellm to our setting, since it requires an LLM call for each observation, where the total number of calls scales to the billions in our case. Therefore, we designed a more scalable variant which i) replaces the LLM embedding with a lightweight bag-of-words embedding \footnote{We also tried using Sentence Transformers (SBERT) for generating sentence embeddings, which led to an approximately 20x throughput drop: 28k vs 1.4k FPS on a Tesla V100-32GB GPU, when the sentence embeddings can be cached. If we recompute the sentence embedding in the forward pass every time, the throughput drop is approximately 100x.} using FastText \citep{bojanowski2016enriching}, and ii) includes an episodic term in the intrinsic reward, as with Motif and \oni (see \Cref{sec:rl-details}). We call it \ellmbow and include more details in Appendix \ref{app:ellm-details}.
We note that \citet{klissarovdoro2023motif} found that other exploration methods based on novelty bonuses such as RND \citep{RND}, NovelD \citep{NovelD} and E3B \citep{E3B} did not improve over the extrinsic reward baseline on these tasks, hence we do not include them. 
\looseness=-1

\subsection{Main Results}
\label{sec:expr_main}
\begin{figure*}[t]
    \centering
    \begin{subfigure}[b]{\linewidth}
        \includegraphics[width=\linewidth]{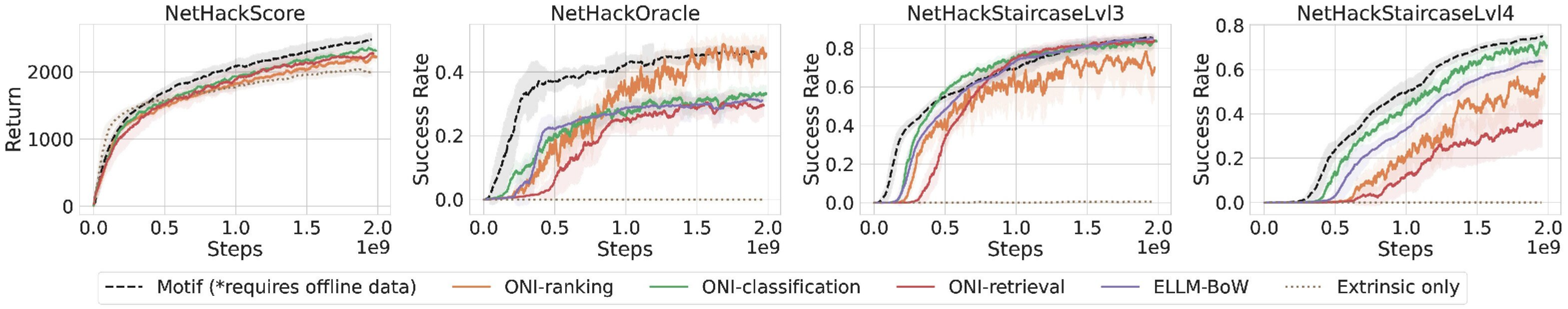}
        \caption{Performance on four different NetHack tasks.}
        \label{fig:main}
    \end{subfigure}
     \begin{subfigure}[b]{\linewidth}
        \includegraphics[width=\linewidth]{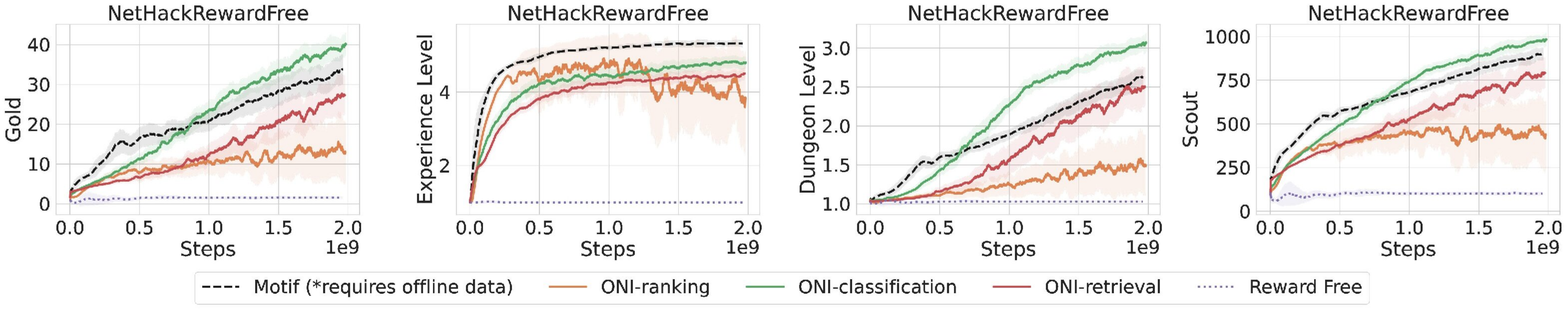}
        \caption{Game progress of intrinsic rewards only agents. }
         \label{fig:main_reward_free}
    \end{subfigure}
    \caption{\oni-based methods are able to match or closely track the performance of Motif without using an pre-collected dataset. This includes (a) reward-based and (b) reward-free settings. Motif's pre-collected dataset uses privileged information about dense reward functions to solve sparse-reward or reward-free environments while \oni-methods do not. \ellmbow demonstrated to be a competitive baseline here too.}
\end{figure*}

\textbf{Task Performance}  We report the average performance
and $95\%$ confidence intervals computed via standard errors over $5$ seeds in \Cref{fig:main}.
The extrinsic reward agent performs reasonably well on the dense \texttt{Score} task, but completely fails on the others due to reward sparsity. We find that, despite not requiring any external data, our \onicls agent is able to match Motif on all tasks except \texttt{Oracle}, where it still performs well. 
We note that Motif requires pre-collecting data from the \texttt{Score} task even for sparse reward tasks---this assumes access to an additional dense reward function, which is an often unrealistic assumption. It also incurs an additional one billion ($10^9$) environment samples prior to policy training.
All of our agents significantly outperform the extrinsic only baseline on the sparse reward tasks, demonstrating they are able to explore effectively while synthesizing their intrinsic rewards from online data alone. \looseness=-1

Despite its simplicity, \oniretrieval performs surprisingly well, and its performance is often close to that of \onicls. This is likely a consequence of many messages with positive valence being repeated in the early game of NetHack, such as ``\emph{You find a hidden passage}'' that allows exploring the rest of the level, or ``\emph{You kill the} \{monster\}!'' which leads to experience gain. \onicls, which also predicts binary rewards but is able to generalize to unseen messages, provides a modest but consistent improvement over \oniretrieval across all environments. This suggests that learning a reward model is indeed helpful. We would expect this gap to increase in settings with added noise or caption diversity, since they increase the likelihood of observed captions being unique. \looseness=-1

We do not observe any significant gains from using \oniranking over \onicls, despite it being more conceptually general and able to represent a continuous range of intrinsic rewards rather than binary values. This may be because our tasks take place at the very earliest part of the game of NetHack, where only a small fraction of all possible messages are observed, which would also explain the relatively strong performance of \oniretrieval. We hypothesize that more benefits will appear in settings with higher observation diversity.%
\ellmbow performs quite well on these tasks, closely tracking or matching Motif and \oni methods. However, in \Cref{sec:comp_cosine_bow} we highlight a fundamental limitation of \ellmbow, namely its inability to capture complex semantic meaning, whereas \oni is capable thanks to its use of an LLM. It is worth noting that directly using \ellmbow as in \citet{du2023guiding} without our episodic bonus completely fails, see \Cref{app:episodic_bonus_cosine_bow}. 

\textbf{Game Progress of Intrinsic-Reward-Only Agents} Results for all methods trained in the reward-free setting are shown in \Cref{fig:main_reward_free}. All of our \oni methods are able to make meaningful progress across all metrics. 
In particular, \onicls outperforms Motif on 3 out of 4 metrics. 
We include the results of Motif for reference, yet emphasize it is actually inapplicable in a truly reward-free setup since it assumes access to dense rewards. %

\subsection{Comparisons for More Complex Goals}
\label{sec:comp_cosine_bow}
\begin{figure}[ht]
    \centering
    \begin{subfigure}[b]{0.49\linewidth}
        \includegraphics[width=0.9\linewidth]{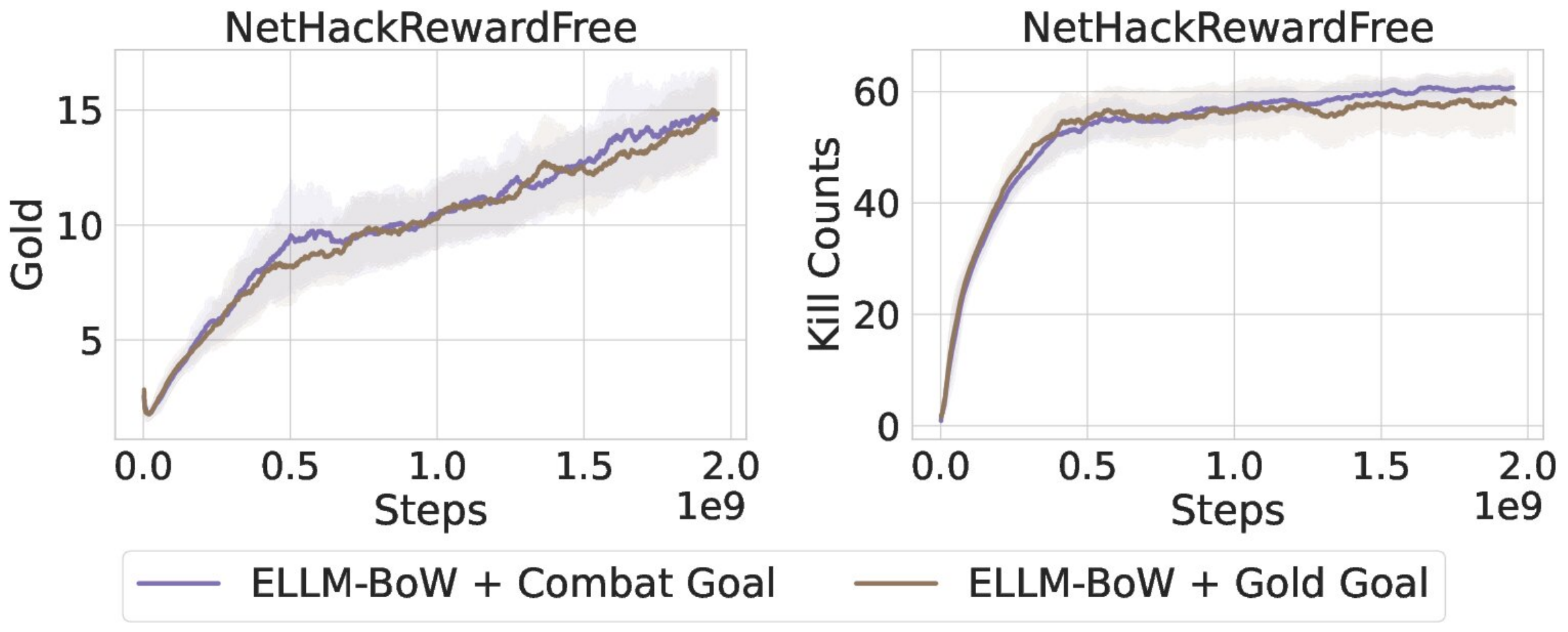}
        \caption{\ellmbow  under 2 different goals}
    \end{subfigure}
    \hfill
    \begin{subfigure}[b]{0.49\linewidth}
        \includegraphics[width=0.9\linewidth]{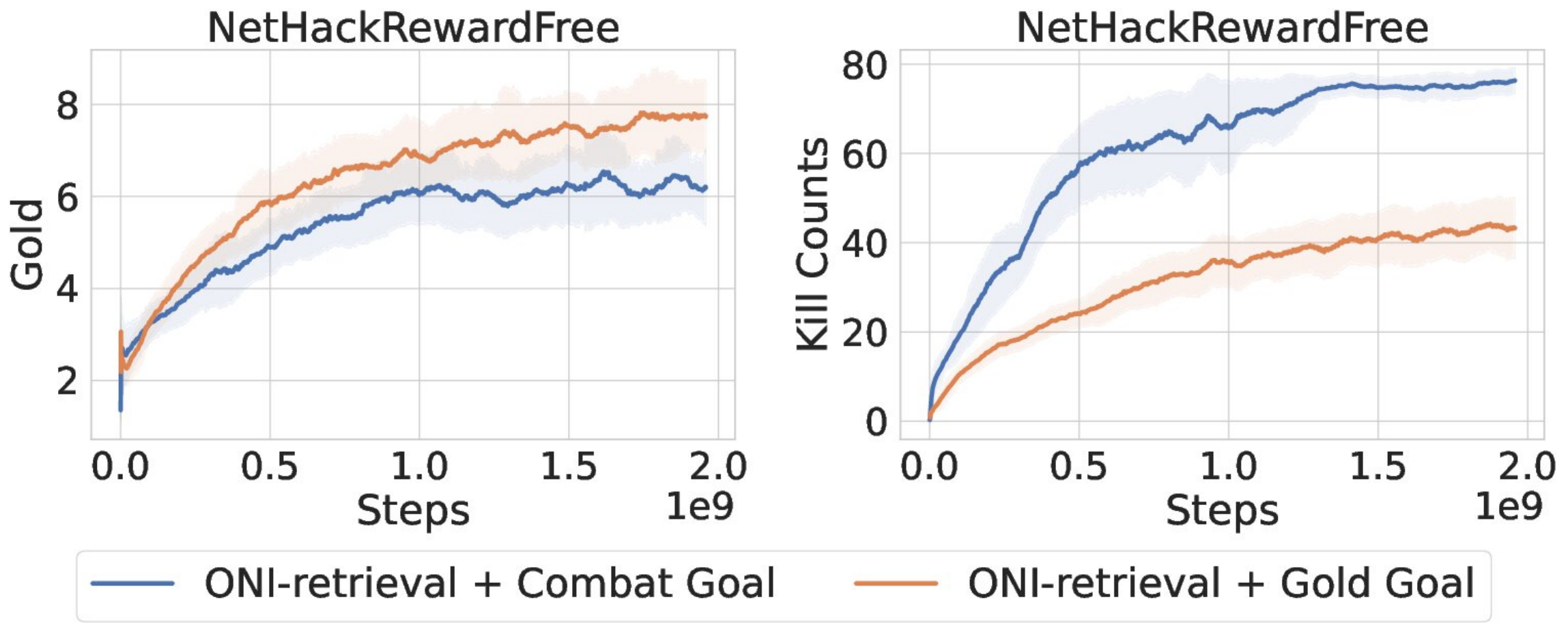} 
        \caption{\onicls  under 2 different goals}
    \end{subfigure}
    \caption{\textbf{(a)} \ellmbow is not able to understand the semantic meaning of complex goals, resulting in agents with similar behavior under the combat and the gold goal. \textbf{(b)} \oniretrieval can distinguish the goals and the resulting agents focus on different aspects of game progress. \looseness=-1 }
    \label{fig:comp_with_cosine_bow_2_goals}
\end{figure}

Even though \ellmbow performs well in \Cref{sec:expr_main}, we have found that it does not capture the semantic meanings of more complex goal strings due to the simple bag-of-word representation. To demonstrate this,
we train \ellmbow and \oniretrieval for two opposite goals in the extrinsic-reward-free environment: \textbf{(Gold)} ``collect gold but do not kill monsters'' vs \textbf{(Combat)} ``kill monsters but do not collect gold''.  \Cref{fig:comp_with_cosine_bow_2_goals} shows that \ellmbow produces two agents of nearly identical behavior in terms of collected gold and killed monsters. In contrast, \oniretrieval is able to distinguish the two goals and produces agents that emphasize either combat engagement or gold collection, depending on the goal string. See \Cref{app:goal_strings} for the goal strings for \ellmbow and prompts for \oniretrieval.

\subsection{Ablation Studies}
\label{sec:ablations}

\begin{figure}[th]
    \centering
    \includegraphics[width=0.9\linewidth]{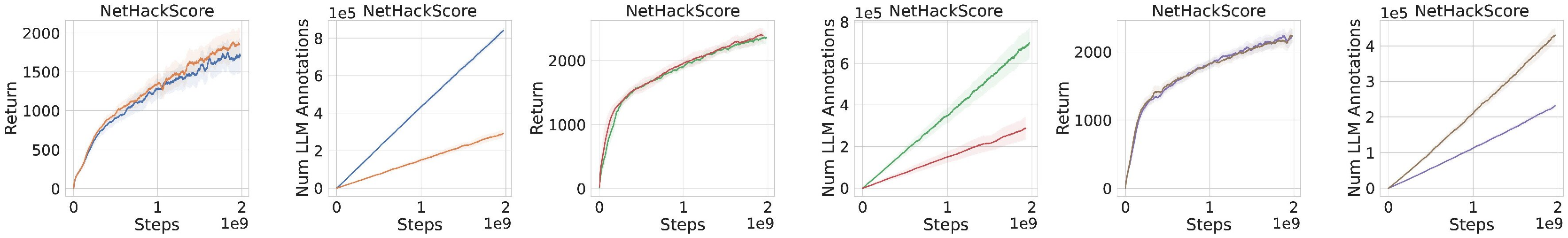}
    \includegraphics[width=0.9\linewidth]{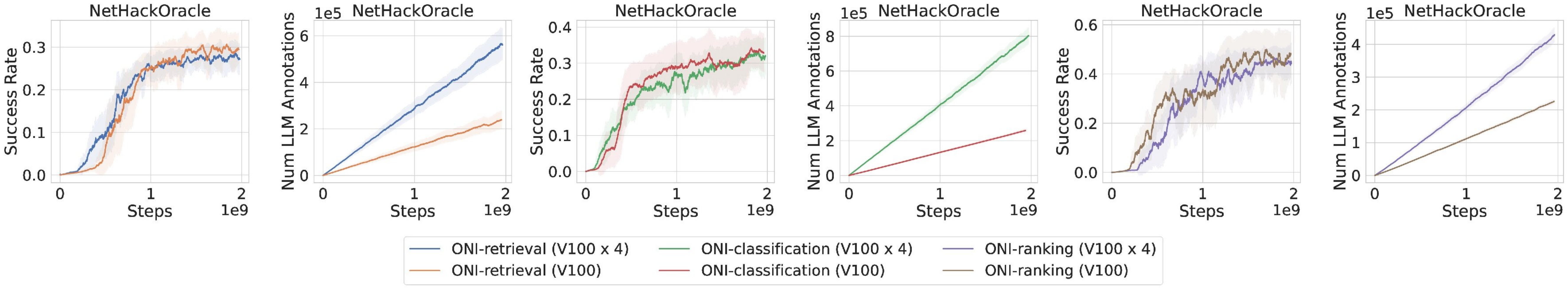}
    \caption{
        Performance remains comparable despite doubling LLM annotation throughput.
        \oniranking's throughput is lower than the others due to annotating pairs of captions.
    }
    \label{fig:4gpu}
\end{figure}

\textbf{Performance vs. LLM Annotation Throughput}
We compared agents trained on the \texttt{Score} and \texttt{Oracle} tasks using either $1$ or $4$ V100 GPUs in the LLM server node. 
As shown in \Cref{fig:4gpu}, using 4 GPUs rather than 1 significantly increases the number of annotated observations, but does not significantly change the performance.
This suggests that many of the labelled examples may contain redundant information, which is not useful for updating the reward model. Designing more sophisticated prioritization schemes, which can select maximally informative examples to send to the LLM, constitutes an interesting direction for future work. 

\begin{figure}[th]
    \centering
    \includegraphics[width=0.43\linewidth]{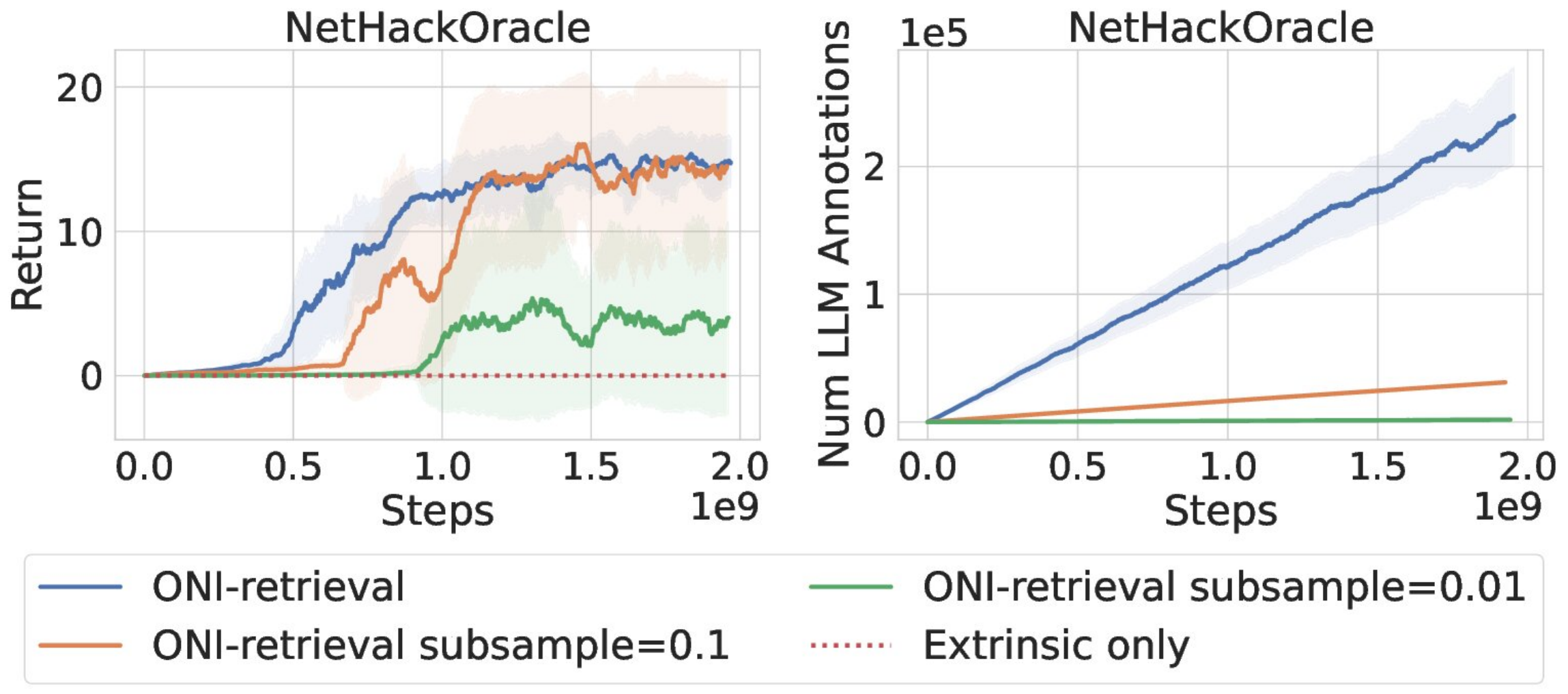}
    \hspace{15pt}
    \includegraphics[width=0.48\linewidth]{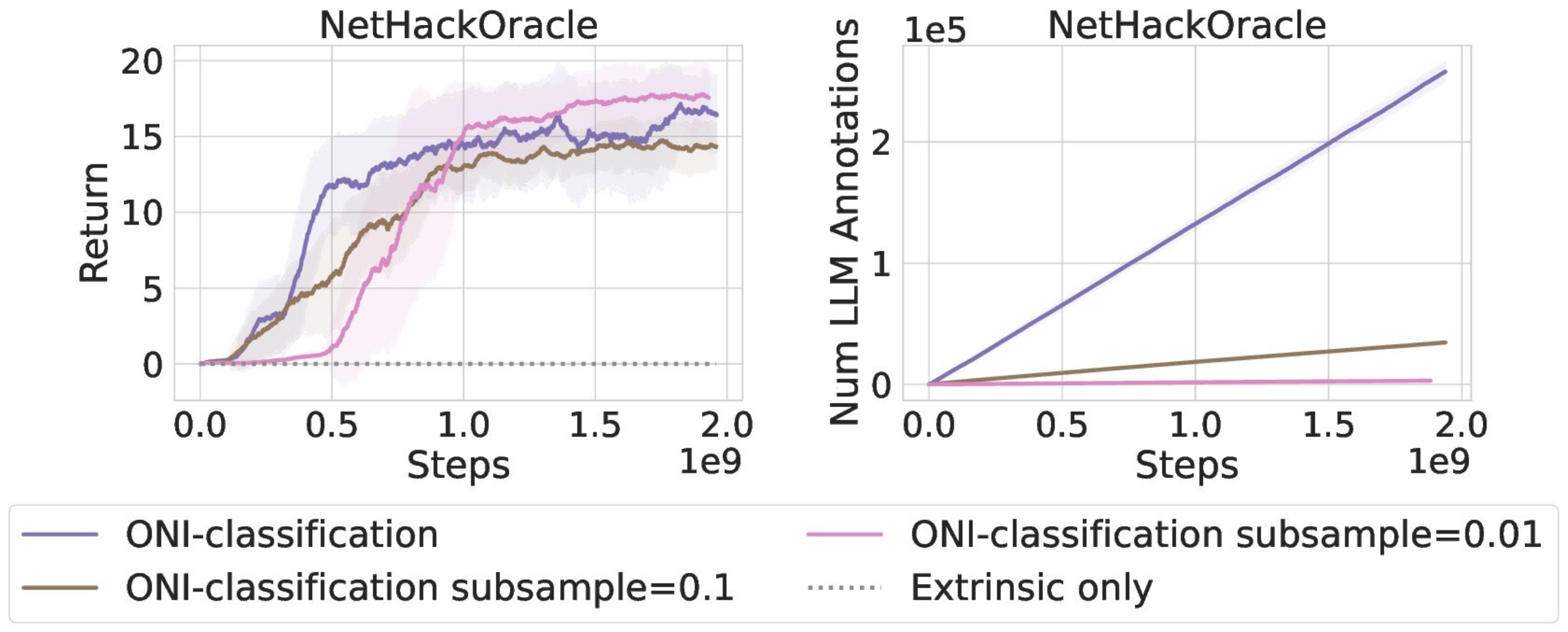}
    \caption{
        Performance of \oniretrieval plunges when the subsampling rate reduces to 0.01, while performance of \onicls is still comparable with the original one. 
    }
    \label{fig:ablation_llm_annotation_subsample}
\end{figure}

\textbf{Performance when Reducing LLM Annotations} Thus far we have been using all LLM annotations available. It is also intriguing to check the performance when we limit the volume of annotations, to simulate more resource-constrained settings. Here we subsample the LLM-annotated messages with rate 0.1 and 0.01 before sending them back to the hash table. \Cref{fig:ablation_llm_annotation_subsample} shows that the performance of \oniretrieval significantly drops with rate 0.01, whereas \onicls remains comparable. This suggests that using a parametric reward model can help reduce the number of annotations required thanks to its generalization ability.

See Appendix \ref{app:additional-ablations} for additional ablations studying hyperparameters and sampling mechanisms.

\section{Conclusion}
We have introduced \oni, a distributed online intrinsic reward and agent learning system. We showed that we are able to match the state of the art across a range of challenging sparse reward tasks from the NetHack Learning Environment, while removing the need for a large pre-collected dataset or auxiliary dense reward function required by previous work. We explored three different instantiations of our system of varying levels of complexity and generality, and study their tradeoffs. Our work paves the way for intrinsic reward methods which can learn purely from agent experience, are not constrained by external dataset size or quality, and can leverage high-performance RL training. \looseness=-1

\bibliography{oni}

\begin{thebibliography}{60}
\providecommand{\natexlab}[1]{#1}
\providecommand{\url}[1]{\texttt{#1}}
\expandafter\ifx\csname urlstyle\endcsname\relax
  \providecommand{\doi}[1]{DOI: #1}\else
  \providecommand{\doi}{DOI: \begingroup \urlstyle{rm}\Url}\fi

\bibitem[Adeniji et~al.(2023)Adeniji, Xie, Sferrazza, Seo, James, and
  Abbeel]{adeniji2023languagerewardmodulationpretraining}
Ademi Adeniji, Amber Xie, Carmelo Sferrazza, Younggyo Seo, Stephen James, and
  Pieter Abbeel.
\newblock Language reward modulation for pretraining reinforcement learning,
  2023.
\newblock URL \url{https://arxiv.org/abs/2308.12270}.

\bibitem[Agarwal et~al.(2020)Agarwal, Henaff, Kakade, and Sun]{PCPG}
Alekh Agarwal, Mikael Henaff, Sham Kakade, and Wen Sun.
\newblock Pc-pg: Policy cover directed exploration for provable policy gradient
  learning.
\newblock \emph{Advances in neural information processing systems},
  33:\penalty0 13399--13412, 2020.

\bibitem[Ahn et~al.(2022)]{ahn2022icanisay}
Michael Ahn et~al.
\newblock Do as i can, not as i say: Grounding language in robotic affordances,
  2022.
\newblock URL \url{https://arxiv.org/abs/2204.01691}.

\bibitem[Bellemare et~al.(2016)Bellemare, Srinivasan, Ostrovski, Schaul,
  Saxton, and Munos]{PseudoCounts}
Marc~G. Bellemare, Sriram Srinivasan, Georg Ostrovski, Tom Schaul, David
  Saxton, and R\'{e}mi Munos.
\newblock Unifying count-based exploration and intrinsic motivation.
\newblock In \emph{Proceedings of the 30th International Conference on Neural
  Information Processing Systems}, NIPS'16, pp.\  1479–1487, Red Hook, NY,
  USA, 2016. Curran Associates Inc.
\newblock ISBN 9781510838819.

\bibitem[Bojanowski et~al.(2016)Bojanowski, Grave, Joulin, and
  Mikolov]{bojanowski2016enriching}
Piotr Bojanowski, Edouard Grave, Armand Joulin, and Tomas Mikolov.
\newblock Enriching word vectors with subword information.
\newblock \emph{arXiv preprint arXiv:1607.04606}, 2016.

\bibitem[Booth et~al.(2023)Booth, Knox, Shah, Niekum, Stone, and
  Allievi]{booth23perils}
Serena Booth, W~Bradley Knox, Julie Shah, Scott Niekum, Peter Stone, and
  Alessandro Allievi.
\newblock The perils of trial-and-error reward design: Misdesign through
  overfitting and invalid task specifications.
\newblock In \emph{Proceedings of the 37th AAAI Conference on Artificial
  Intelligence (AAAI)}, Feb 2023.

\bibitem[Bradley \& Terry(1952)Bradley and Terry]{bradley1952rank}
Ralph~Allan Bradley and Milton~E Terry.
\newblock Rank analysis of incomplete block designs: I. the method of paired
  comparisons.
\newblock \emph{Biometrika}, 39\penalty0 (3/4):\penalty0 324--345, 1952.

\bibitem[Brafman \& Tennenholtz(2002)Brafman and Tennenholtz]{Rmax}
Ronen~I. Brafman and Moshe Tennenholtz.
\newblock {R-MAX} - {A} general polynomial time algorithm for near-optimal
  reinforcement learning.
\newblock \emph{J. Mach. Learn. Res.}, 3:\penalty0 213--231, 2002.

\bibitem[Burda et~al.(2019)Burda, Edwards, Storkey, and Klimov]{RND}
Yuri Burda, Harrison Edwards, Amos Storkey, and Oleg Klimov.
\newblock Exploration by random network distillation.
\newblock In \emph{International Conference on Learning Representations}, 2019.

\bibitem[Cao et~al.(2024)Cao, Zhao, Cheng, Shu, Liu, Liang, Zhao, and
  Li]{cao2024survey}
Yuji Cao, Huan Zhao, Yuheng Cheng, Ting Shu, Guolong Liu, Gaoqi Liang, Junhua
  Zhao, and Yun Li.
\newblock Survey on large language model-enhanced reinforcement learning:
  Concept, taxonomy, and methods.
\newblock \emph{arXiv preprint arXiv:2404.00282}, 2024.

\bibitem[Carta et~al.(2022)Carta, Oudeyer, Sigaud, and
  Lamprier]{carta2022eager}
Thomas Carta, Pierre-Yves Oudeyer, Olivier Sigaud, and Sylvain Lamprier.
\newblock Eager: Asking and answering questions for automatic reward shaping in
  language-guided rl.
\newblock \emph{Advances in Neural Information Processing Systems},
  35:\penalty0 12478--12490, 2022.

\bibitem[Cho et~al.(2014)Cho, van Merrienboer, G{\"{u}}l{\c{c}}ehre, Bougares,
  Schwenk, and Bengio]{GRU}
Kyunghyun Cho, Bart van Merrienboer, {\c{C}}aglar G{\"{u}}l{\c{c}}ehre, Fethi
  Bougares, Holger Schwenk, and Yoshua Bengio.
\newblock Learning phrase representations using {RNN} encoder-decoder for
  statistical machine translation.
\newblock \emph{CoRR}, abs/1406.1078, 2014.
\newblock URL \url{http://arxiv.org/abs/1406.1078}.

\bibitem[Chu et~al.(2023)Chu, Zhao, Weber, Li, and
  Wermter]{chu2023accelerating}
Kun Chu, Xufeng Zhao, Cornelius Weber, Mengdi Li, and Stefan Wermter.
\newblock Accelerating reinforcement learning of robotic manipulations via
  feedback from large language models.
\newblock \emph{arXiv preprint arXiv:2311.02379}, 2023.

\bibitem[Clevert et~al.(2016)Clevert, Unterthiner, and Hochreiter]{ELU}
Djork-Arné Clevert, Thomas Unterthiner, and Sepp Hochreiter.
\newblock Fast and accurate deep network learning by exponential linear units
  (elus).
\newblock In Yoshua Bengio and Yann LeCun (eds.), \emph{ICLR}, 2016.
\newblock URL
  \url{http://dblp.uni-trier.de/db/conf/iclr/iclr2016.html#ClevertUH15}.

\bibitem[Driess et~al.(2023)Driess, , et~al.]{driess2023palme}
Danny Driess, , et~al.
\newblock Palm-e: An embodied multimodal language model.
\newblock In \emph{arXiv preprint arXiv:2303.03378}, 2023.

\bibitem[Du et~al.(2023{\natexlab{a}})Du, Watkins, Wang, Colas, Darrell,
  Abbeel, Gupta, and Andreas]{du2023guiding}
Yuqing Du, Olivia Watkins, Zihan Wang, C{\'e}dric Colas, Trevor Darrell, Pieter
  Abbeel, Abhishek Gupta, and Jacob Andreas.
\newblock Guiding pretraining in reinforcement learning with large language
  models.
\newblock In \emph{International Conference on Machine Learning}, pp.\
  8657--8677. PMLR, 2023{\natexlab{a}}.

\bibitem[Du et~al.(2023{\natexlab{b}})Du, Watkins, Wang, Colas, Darrell,
  Abbeel, Gupta, and Andreas]{pmlr-v202-du23f}
Yuqing Du, Olivia Watkins, Zihan Wang, C\'{e}dric Colas, Trevor Darrell, Pieter
  Abbeel, Abhishek Gupta, and Jacob Andreas.
\newblock Guiding pretraining in reinforcement learning with large language
  models.
\newblock In Andreas Krause, Emma Brunskill, Kyunghyun Cho, Barbara Engelhardt,
  Sivan Sabato, and Jonathan Scarlett (eds.), \emph{Proceedings of the 40th
  International Conference on Machine Learning}, volume 202 of
  \emph{Proceedings of Machine Learning Research}, pp.\  8657--8677. PMLR,
  23--29 Jul 2023{\natexlab{b}}.
\newblock URL \url{https://proceedings.mlr.press/v202/du23f.html}.

\bibitem[Dubey et~al.(2024)]{dubey2024llama3herdmodels}
Abhimanyu Dubey et~al.
\newblock The llama 3 herd of models, 2024.
\newblock URL \url{https://arxiv.org/abs/2407.21783}.

\bibitem[Ecoffet et~al.(2019)Ecoffet, Huizinga, Lehman, Stanley, and
  Clune]{ecoffet2019go}
Adrien Ecoffet, Joost Huizinga, Joel Lehman, Kenneth~O Stanley, and Jeff Clune.
\newblock Go-explore: a new approach for hard-exploration problems.
\newblock \emph{arXiv preprint arXiv:1901.10995}, 2019.

\bibitem[Fan et~al.(2022)Fan, Wang, Jiang, Mandlekar, Yang, Zhu, Tang, Huang,
  Zhu, and Anandkumar]{fan2022minedojo}
Linxi Fan, Guanzhi Wang, Yunfan Jiang, Ajay Mandlekar, Yuncong Yang, Haoyi Zhu,
  Andrew Tang, De-An Huang, Yuke Zhu, and Anima Anandkumar.
\newblock Minedojo: Building open-ended embodied agents with internet-scale
  knowledge.
\newblock In \emph{Thirty-sixth Conference on Neural Information Processing
  Systems Datasets and Benchmarks Track}, 2022.
\newblock URL \url{https://openreview.net/forum?id=rc8o_j8I8PX}.

\bibitem[Gomez et~al.(2024)Gomez, Bowling, and
  Machado]{gomez2024properlaplacianrepresentationlearning}
Diego Gomez, Michael Bowling, and Marlos~C. Machado.
\newblock Proper laplacian representation learning, 2024.
\newblock URL \url{https://arxiv.org/abs/2310.10833}.

\bibitem[Henaff et~al.(2022{\natexlab{a}})Henaff, Raileanu, Jiang, and
  Rockt{\"a}schel]{E3B}
Mikael Henaff, Roberta Raileanu, Minqi Jiang, and Tim Rockt{\"a}schel.
\newblock Exploration via elliptical episodic bonuses.
\newblock In Alice~H. Oh, Alekh Agarwal, Danielle Belgrave, and Kyunghyun Cho
  (eds.), \emph{Advances in Neural Information Processing Systems},
  2022{\natexlab{a}}.

\bibitem[Henaff et~al.(2022{\natexlab{b}})Henaff, Raileanu, Jiang, and
  Rockt{\"a}schel]{henaff2022exploration}
Mikael Henaff, Roberta Raileanu, Minqi Jiang, and Tim Rockt{\"a}schel.
\newblock Exploration via elliptical episodic bonuses.
\newblock \emph{Advances in Neural Information Processing Systems},
  35:\penalty0 37631--37646, 2022{\natexlab{b}}.

\bibitem[Ibrahim et~al.(2024)Ibrahim, Mostafa, Jnadi, and
  Osinenko]{ibrahim2024comprehensive}
Sinan Ibrahim, Mostafa Mostafa, Ali Jnadi, and Pavel Osinenko.
\newblock Comprehensive overview of reward engineering and shaping in advancing
  reinforcement learning applications.
\newblock \emph{arXiv preprint arXiv:2408.10215}, 2024.

\bibitem[Jeurissen et~al.(2024)Jeurissen, Perez-Liebana, Gow, Cakmak, and
  Kwan]{jeurissen2024playing}
Dominik Jeurissen, Diego Perez-Liebana, Jeremy Gow, Duygu Cakmak, and James
  Kwan.
\newblock Playing nethack with llms: Potential \& limitations as zero-shot
  agents.
\newblock \emph{arXiv preprint arXiv:2403.00690}, 2024.

\bibitem[Kearns \& Singh(2002)Kearns and Singh]{E3}
Michael Kearns and Satinder Singh.
\newblock Near-optimal reinforcement learning in polynomial time.
\newblock In \emph{Machine Learning}, pp.\  209--232. Morgan Kaufmann, 2002.

\bibitem[Kim et~al.(2024)Kim, Seo, Liu, Lee, Shin, Lee, and Lee]{kim2024guide}
Changyeon Kim, Younggyo Seo, Hao Liu, Lisa Lee, Jinwoo Shin, Honglak Lee, and
  Kimin Lee.
\newblock Guide your agent with adaptive multimodal rewards.
\newblock \emph{Advances in Neural Information Processing Systems}, 36, 2024.

\bibitem[Kingma(2014)]{kingma2014adam}
Diederik~P Kingma.
\newblock Adam: A method for stochastic optimization.
\newblock \emph{arXiv preprint arXiv:1412.6980}, 2014.

\bibitem[Klissarov et~al.(2023)Klissarov, D’Oro, Sodhani, Raileanu, Bacon,
  Vincent, Zhang, and Henaff]{klissarovdoro2023motif}
Martin Klissarov, Pierluca D’Oro, Shagun Sodhani, Roberta Raileanu,
  Pierre-Luc Bacon, Pascal Vincent, Amy Zhang, and Mikael Henaff.
\newblock Motif: Intrinsic motivation from artificial intelligence feedback.
\newblock \emph{arXiv preprint arXiv:2310.00166}, 9 2023.

\bibitem[Klissarov et~al.(2025)Klissarov, Henaff, Raileanu, Sodhani, Vincent,
  Zhang, Bacon, Precup, Machado, and D'Oro]{klissarov2024maestromotif}
Martin Klissarov, Mikael Henaff, Roberta Raileanu, Shagun Sodhani, Pascal
  Vincent, Amy Zhang, Pierre-Luc Bacon, Doina Precup, Marlos~C. Machado, and
  Pierluca D'Oro.
\newblock Maestromotif: Skill design from artificial intelligence feedback.
\newblock 2025.
\newblock URL \url{https://openreview.net/forum?id=or8mMhmyRV}.

\bibitem[Kurenkov et~al.(2023)Kurenkov, Nikulin, Tarasov, and
  Kolesnikov]{kurenkov2023katakomba}
Vladislav Kurenkov, Alexander Nikulin, Denis Tarasov, and Sergey Kolesnikov.
\newblock Katakomba: Tools and benchmarks for data-driven nethack, 2023.

\bibitem[K{\"{u}}ttler et~al.(2020)K{\"{u}}ttler, Nardelli, Miller, Raileanu,
  Selvatici, Grefenstette, and Rockt{\"{a}}schel]{kuettler2020nethack}
Heinrich K{\"{u}}ttler, Nantas Nardelli, Alexander~H. Miller, Roberta Raileanu,
  Marco Selvatici, Edward Grefenstette, and Tim Rockt{\"{a}}schel.
\newblock {The NetHack Learning Environment}.
\newblock In \emph{Proceedings of the Conference on Neural Information
  Processing Systems (NeurIPS)}, 2020.

\bibitem[Kwon et~al.(2023{\natexlab{a}})Kwon, Xie, Bullard, and
  Sadigh]{kwon2023reward}
Minae Kwon, Sang~Michael Xie, Kalesha Bullard, and Dorsa Sadigh.
\newblock Reward design with language models.
\newblock In \emph{The Eleventh International Conference on Learning
  Representations}, 2023{\natexlab{a}}.
\newblock URL \url{https://openreview.net/forum?id=10uNUgI5Kl}.

\bibitem[Kwon et~al.(2023{\natexlab{b}})Kwon, Li, Zhuang, Sheng, Zheng, Yu,
  Gonzalez, Zhang, and Stoica]{kwon2023efficient}
Woosuk Kwon, Zhuohan Li, Siyuan Zhuang, Ying Sheng, Lianmin Zheng, Cody~Hao Yu,
  Joseph~E. Gonzalez, Hao Zhang, and Ion Stoica.
\newblock Efficient memory management for large language model serving with
  pagedattention.
\newblock In \emph{Proceedings of the ACM SIGOPS 29th Symposium on Operating
  Systems Principles}, 2023{\natexlab{b}}.

\bibitem[Li et~al.(2024)Li, Yang, Wang, Zhu, Zhou, Qiao, Wang, Li, Lu, and
  Dai]{li2024auto}
Hao Li, Xue Yang, Zhaokai Wang, Xizhou Zhu, Jie Zhou, Yu~Qiao, Xiaogang Wang,
  Hongsheng Li, Lewei Lu, and Jifeng Dai.
\newblock Auto mc-reward: Automated dense reward design with large language
  models for minecraft.
\newblock In \emph{Proceedings of the IEEE/CVF Conference on Computer Vision
  and Pattern Recognition}, pp.\  16426--16435, 2024.

\bibitem[Lu et~al.(2024)Lu, Hu, and Clune]{lu2024intelligent}
Cong Lu, Shengran Hu, and Jeff Clune.
\newblock Intelligent go-explore: Standing on the shoulders of giant foundation
  models.
\newblock \emph{arXiv preprint arXiv:2405.15143}, 2024.

\bibitem[Ma et~al.(2023)Ma, Liang, Wang, Huang, Bastani, Jayaraman, Zhu, Fan,
  and Anandkumar]{ma2023eureka}
Yecheng~Jason Ma, William Liang, Guanzhi Wang, De-An Huang, Osbert Bastani,
  Dinesh Jayaraman, Yuke Zhu, Linxi Fan, and Anima Anandkumar.
\newblock Eureka: Human-level reward design via coding large language models.
\newblock \emph{arXiv preprint arXiv: Arxiv-2310.12931}, 2023.

\bibitem[Myffili(2021)]{CDGPT5}
Anssi Myffili.
\newblock Nle challenge baseline using sample-factory.
\newblock \url{https://github.com/Miffyli/nle-sample-factory-baseline}, 2021.

\bibitem[Ng et~al.(1999)Ng, Harada, and Russell]{ng1999rewardshaping}
Andrew~Y. Ng, Daishi Harada, and Stuart~J. Russell.
\newblock Policy invariance under reward transformations: Theory and
  application to reward shaping.
\newblock In \emph{Proceedings of the Sixteenth International Conference on
  Machine Learning}, ICML '99, pp.\  278–287, San Francisco, CA, USA, 1999.
  Morgan Kaufmann Publishers Inc.
\newblock ISBN 1558606122.

\bibitem[Pathak et~al.(2017)Pathak, Agrawal, Efros, and Darrell]{ICM}
Deepak Pathak, Pulkit Agrawal, Alexei~A. Efros, and Trevor Darrell.
\newblock Curiosity-driven exploration by self-supervised prediction.
\newblock \emph{CoRR}, abs/1705.05363, 2017.

\bibitem[Petrenko et~al.(2020)Petrenko, Huang, Kumar, Sukhatme, and
  Koltun]{petrenko2020sf}
Aleksei Petrenko, Zhehui Huang, Tushar Kumar, Gaurav~S. Sukhatme, and Vladlen
  Koltun.
\newblock Sample factory: Egocentric 3d control from pixels at 100000 {FPS}
  with asynchronous reinforcement learning.
\newblock In \emph{Proceedings of the 37th International Conference on Machine
  Learning, {ICML} 2020, 13-18 July 2020, Virtual Event}, volume 119 of
  \emph{Proceedings of Machine Learning Research}, pp.\  7652--7662. {PMLR},
  2020.
\newblock URL \url{http://proceedings.mlr.press/v119/petrenko20a.html}.

\bibitem[Piterbarg et~al.(2023)Piterbarg, Pinto, and
  Fergus]{piterbarg2023nethack}
Ulyana Piterbarg, Lerrel Pinto, and Rob Fergus.
\newblock Nethack is hard to hack.
\newblock In \emph{Thirty-seventh Conference on Neural Information Processing
  Systems}, 2023.
\newblock URL \url{https://openreview.net/forum?id=tp2nEZ5zfP}.

\bibitem[Raileanu \& Rocktäschel(2020)Raileanu and Rocktäschel]{RIDE}
Roberta Raileanu and Tim Rocktäschel.
\newblock Ride: Rewarding impact-driven exploration for procedurally-generated
  environments.
\newblock In \emph{International Conference on Learning Representations}, 2020.

\bibitem[Randlov \& Alstrøm(1998)Randlov and Alstrøm]{randlov1998shaping}
Jette Randlov and Preben Alstrøm.
\newblock Learning to drive a bicycle using reinforcement learning and shaping.
\newblock pp.\  463--471, 01 1998.

\bibitem[Rocamonde et~al.(2023)Rocamonde, Montesinos, Nava, Perez, and
  Lindner]{rocamonde2023vision}
Juan Rocamonde, Victoriano Montesinos, Elvis Nava, Ethan Perez, and David
  Lindner.
\newblock Vision-language models are zero-shot reward models for reinforcement
  learning.
\newblock \emph{arXiv preprint arXiv:2310.12921}, 2023.

\bibitem[Schmidhuber(1991)]{schmidhuber1991possibility}
J{\"u}rgen Schmidhuber.
\newblock A possibility for implementing curiosity and boredom in
  model-building neural controllers.
\newblock In \emph{Proc. of the international conference on simulation of
  adaptive behavior: From animals to animats}, pp.\  222--227, 1991.

\bibitem[Schulman et~al.(2017)Schulman, Wolski, Dhariwal, Radford, and
  Klimov]{schulman2017proximal}
John Schulman, Filip Wolski, Prafulla Dhariwal, Alec Radford, and Oleg Klimov.
\newblock Proximal policy optimization algorithms.
\newblock \emph{arXiv preprint arXiv:1707.06347}, 2017.

\bibitem[Shyam et~al.(2019)Shyam, Ja{\'{s}}kowski, and Gomez]{max}
Pranav Shyam, Wojciech Ja{\'{s}}kowski, and Faustino Gomez.
\newblock Model-based active exploration.
\newblock In Kamalika Chaudhuri and Ruslan Salakhutdinov (eds.),
  \emph{Proceedings of the 36th International Conference on Machine Learning},
  volume~97 of \emph{Proceedings of Machine Learning Research}, pp.\
  5779--5788. PMLR, 09--15 Jun 2019.

\bibitem[Singh et~al.(2010)Singh, Lewis, Barto, and
  Sorg]{singh2010intrinsically}
Satinder Singh, Richard~L Lewis, Andrew~G Barto, and Jonathan Sorg.
\newblock Intrinsically motivated reinforcement learning: An evolutionary
  perspective.
\newblock \emph{IEEE Transactions on Autonomous Mental Development}, 2\penalty0
  (2):\penalty0 70--82, 2010.

\bibitem[Sorg et~al.(2010)Sorg, Singh, and Lewis]{sorg2010internal}
Jonathan Sorg, Satinder~P Singh, and Richard~L Lewis.
\newblock Internal rewards mitigate agent boundedness.
\newblock In \emph{Proceedings of the 27th international conference on machine
  learning (ICML-10)}, pp.\  1007--1014, 2010.

\bibitem[Stadie et~al.(2015)Stadie, Levine, and
  Abbeel]{stadie2015incentivizing}
Bradly~C Stadie, Sergey Levine, and Pieter Abbeel.
\newblock Incentivizing exploration in reinforcement learning with deep
  predictive models.
\newblock \emph{arXiv preprint arXiv:1507.00814}, 2015.

\bibitem[Sutton \& Barto(2018)Sutton and Barto]{Sutton1998}
Richard~S. Sutton and Andrew~G. Barto.
\newblock \emph{Reinforcement Learning: An Introduction}.
\newblock The MIT Press, second edition, 2018.
\newblock URL \url{http://incompleteideas.net/book/the-book-2nd.html}.

\bibitem[Wang et~al.(2024)Wang, Xie, Jiang, Mandlekar, Xiao, Zhu, Fan, and
  Anandkumar]{wang2024voyager}
Guanzhi Wang, Yuqi Xie, Yunfan Jiang, Ajay Mandlekar, Chaowei Xiao, Yuke Zhu,
  Linxi Fan, and Anima Anandkumar.
\newblock Voyager: An open-ended embodied agent with large language models.
\newblock \emph{Transactions on Machine Learning Research}, 2024.
\newblock ISSN 2835-8856.
\newblock URL \url{https://openreview.net/forum?id=ehfRiF0R3a}.

\bibitem[Wang et~al.(2021)Wang, Zhou, Zhang, Shao, Hooi, and
  Feng]{DBLP:journals/corr/abs-2107-05545}
Kaixin Wang, Kuangqi Zhou, Qixin Zhang, Jie Shao, Bryan Hooi, and Jiashi Feng.
\newblock Towards better laplacian representation in reinforcement learning
  with generalized graph drawing.
\newblock \emph{CoRR}, abs/2107.05545, 2021.
\newblock URL \url{https://arxiv.org/abs/2107.05545}.

\bibitem[Wu et~al.(2019)Wu, Tucker, and Nachum]{wu2018the}
Yifan Wu, George Tucker, and Ofir Nachum.
\newblock The laplacian in {RL}: Learning representations with efficient
  approximations.
\newblock In \emph{International Conference on Learning Representations}, 2019.
\newblock URL \url{https://openreview.net/forum?id=HJlNpoA5YQ}.

\bibitem[Wu et~al.(2024)Wu, Fan, Liang, Azaria, Li, and Mitchell]{wu2024read}
Yue Wu, Yewen Fan, Paul~Pu Liang, Amos Azaria, Yuanzhi Li, and Tom~M Mitchell.
\newblock Read and reap the rewards: Learning to play atari with the help of
  instruction manuals.
\newblock \emph{Advances in Neural Information Processing Systems}, 36, 2024.

\bibitem[Xie et~al.(2023)Xie, Zhao, Wu, Liu, Luo, Zhong, Yang, and
  Yu]{xie2023text2reward}
Tianbao Xie, Siheng Zhao, Chen~Henry Wu, Yitao Liu, Qian Luo, Victor Zhong,
  Yanchao Yang, and Tao Yu.
\newblock Text2reward: Automated dense reward function generation for
  reinforcement learning.
\newblock \emph{arXiv preprint arXiv:2309.11489}, 2023.

\bibitem[Yu et~al.(2023)Yu, Gileadi, Fu, Kirmani, Lee, Arenas, Chiang, Erez,
  Hasenclever, Humplik, et~al.]{yu2023language}
Wenhao Yu, Nimrod Gileadi, Chuyuan Fu, Sean Kirmani, Kuang-Huei Lee,
  Montse~Gonzalez Arenas, Hao-Tien~Lewis Chiang, Tom Erez, Leonard Hasenclever,
  Jan Humplik, et~al.
\newblock Language to rewards for robotic skill synthesis.
\newblock \emph{arXiv preprint arXiv:2306.08647}, 2023.

\bibitem[Zhang et~al.(2021)Zhang, Xu, Wang, Wu, Keutzer, Gonzalez, and
  Tian]{NovelD}
Tianjun Zhang, Huazhe Xu, Xiaolong Wang, Yi~Wu, Kurt Keutzer, Joseph~E.
  Gonzalez, and Yuandong Tian.
\newblock Noveld: A simple yet effective exploration criterion.
\newblock In A.~Beygelzimer, Y.~Dauphin, P.~Liang, and J.~Wortman Vaughan
  (eds.), \emph{Advances in Neural Information Processing Systems}, 2021.

\bibitem[Zhang et~al.(2015)Zhang, Zhao, and
  LeCun]{DBLP:journals/corr/ZhangZL15}
Xiang Zhang, Junbo~Jake Zhao, and Yann LeCun.
\newblock Character-level convolutional networks for text classification.
\newblock \emph{CoRR}, abs/1509.01626, 2015.
\newblock URL \url{http://arxiv.org/abs/1509.01626}.

\end{thebibliography}
\bibliographystyle{rlj}

\clearpage
\onecolumn
\appendix
\section{Experimental Details}
\label{sec:exp-details}

\subsection{Architectures}
\label{sec:architecture-details}

Our architectures largely follow those used in \citet{klissarovdoro2023motif}. 
\paragraph{Our policy network} uses the Chaotic Dwarven GPT5 architecture originally introduced in \citet{CDGPT5}. This architecture combines convolutional layers processing the top-down visible map centered at the agent with fully-connected layers processing messages and bottom-line statistics including hit points, experience, hunger level and the like. The convolutional encoder has $3$ convolutional layers with $32, 64, 128$ feature maps respectively, interleaved with exponential linear unit (ELU) non-linearities \citep{ELU}.  Messages and bottom-line statistics are each processed with 2-layer MLPs with $128$ hidden units each. All embeddings are combined, passed through a single-layer MLP with $512$ hidden units, and then passed to a recurrent GRU module \citep{GRU} with $512$ hidden units. Finally, this hidden representation feeds into linear critic and actor heads. 

\paragraph{Our reward model} The reward model of \oni-ranking is based on the encoder from \citet{kuettler2020nethack} that processes both state representation and messages. Messages are processed by a $5$-layer character-level CNN \citep{DBLP:journals/corr/ZhangZL15} with $64$ feature maps at each layer. The first, second and last layers are interleaved with max-pooling layers with kernel size and stride $3$. The output is then passed through a $3$-layer MLP with $128, 256, 512$ hidden units at each layer respectively, and ReLU non-linearities, followed by a scalar output. The reward model of \oni-classification only process messages, using the same architecture described above.

\subsection{Hyperparameters}
\label{sec:rl-details}
Following \citet{klissarovdoro2023motif}, we scale the environment reward by 0.1 for the \texttt{Score} task and by 10 for the other sparse reward tasks, and use normalized intrinsic reward
\begin{equation}
    \label{eq:episodic_normalization}
    \rint_{\text{normalized}}(o_t) = \rint(o_t) / N(c_t)^z,
\end{equation}
where $N(c_t)$ is the number of times the caption $c_t$ has been found in one episode. For all our experiments, we use $z=3.0$. This is also called the \emph{episodic bonus}~\cite{henaff2022exploration}.

For \onicls and \oniranking, we train the reward model using the Adam optimizer~\cite{kingma2014adam} with batch size $256$.
\onicls is trained with learning rate $0.0001$, classification threshold $\eta=0.5$, $\beta=0.1$ for the \texttt{Score} task and $\beta=0.4$ for the others.
\oniranking is trained with $0.00001$, $\beta=0.05$ and $\nu_\N = 0$ (mean of the standard normal distribution).
\oniretrieval does not train a reward model, and we use $\beta=0.1$ for the \texttt{Score} task and $\beta=0.5$ for the others.

\Cref{tab:common-appo-hps} shows the APPO hyperparameters which are common to all experiments and \Cref{tab:llm-hps} includes the hyperparameters for the online LLM annotation.
\begin{table}[H]
    \centering
    \begin{tabular}{ll}
    Hyperparameter & Value \\
    \midrule
    Number of Parallel Environment Instances & $480$ \\
    Batch Size & $4096$ \\
    PPO Clip Ratio & $0.1$ \\
    PPO Clip Value & $1.0$ \\
    PPO Epochs & $1$ \\
    Max Gradient Norm & $4.0$ \\
    Value Loss Coefficient & $0.5$ \\
    Exploration Loss & entropy \\
    \end{tabular}
    \caption{Common APPO hyperparameters across all experiments.}
    \label{tab:common-appo-hps}
\end{table}

\begin{table}[H]
    \centering
    \begin{tabular}{ll}
    Hyperparameter & Value \\
    \midrule
    LLM Model & LLaMA-3.1-8B-Instruct \\
    Temperature & $0.1$ \\
    Max tokens to generate & $4096$ \\
    Annotation Batch Size  & $100$ \\
    \end{tabular}
    \caption{LLM annotation hyperparameters.}
    \label{tab:llm-hps}
\end{table}

\subsection{Implementation Details for \ellmbow}
\label{app:ellm-details}

For any piece of text $c$, we construct an embedding vector by first applying the FastText tokenizer \cite{bojanowski2016enriching} to the text, separating it into words $c = (w_1, ..., w_m)$. We then retrieve the FastText word embeddings $x_1,...,x_m$ for each word, and compute their sum $\texttt{embed}(c) = \sum_{i=1}^m x_i$. The intrinsic reward is then defined by the cosine distance between the embeddings of the caption of the current observation and the goal, weighted by an episodic count-based term::

\begin{equation*}
    r(o_t) = \frac{1}{N_e(c_t)^z} \cdot \frac{\texttt{embed}(c_t) \cdot \texttt{embed}(g)}{\|\texttt{embed}(c_t)\| \cdot \|\texttt{embed}(g)\|}
\end{equation*}

We include the episodic term for consistency with Motif and \oni. In Section \ref{app:episodic_bonus_cosine_bow}, we ablate this choice and show that it is essential for good performance. 

\newpage
\section{Prompts}
\label{app:prompts}
Here we include all the LLM prompts we use. The system prompt and the prompt for \oni-ranking are identical to the ones used in \citet{klissarovdoro2023motif}. We adapted these to produce labels for single observations for \oni-retrieval and \oni-classification.   

\textbf{System Prompt (used by all methods)}

\begin{lstlisting}
You are a helpful and honest judge of good gameplaying and progress in the NetHack game. Always answer as helpfully as possible, while being truthful.

If you don't know the answer to a question, please don't share false information.
\end{lstlisting}

\textbf{Prompt for \oni-retrieval and \oni-classification}

\begin{lstlisting}
Context: I am playing NetHack and I will present you a message from the gameplay.

Instruction: First, tell me about your knowledge of NetHack. Mention the goal of NetHack.
Then, write an analysis describing the semantics of each message using your knowledge of NetHack. Finally, respond by assigning a score to the message, which measures whether the status represented by the message is helpful or unhelpful to make some progress towards the goal.
Helpful here does not mean whether it is informative to the player to make decisions or not, but whether it is helpful for the agent to solve the game.
I prefer agents that maximize the score in the game, for instance by killing monsters, collecting gold or going down the stairs in the dungeon.
For extremely helpful messages, assign a label FOO. For unhelpful, neutral, and lukewarm messages, assign a label BAR.

Example 1: "You kill the goblin!  Welcome to experience level 4" indicates that the player
has killed a goblin and gained experience. This led to a level up and increased abilities.
Gaining experience and leveling up are both extremely helpful and this message should be labelled as FOO.

Example 2: "You hear someone counting money." indicates that the player is close to a source of gold, as the sound of counting money suggests a nearby NPC or monster with treasure.
This is also extremely helpful and should be labelled as FOO.

Example 3: "5 gold pieces." indicates you collected 5 gold coins. This is extremely helpful should be labelled as FOO.

Example 4: "That door is closed." is neutral and should be labelled as BAR.

Please rate this message: {{"{}"}}

Constraint: Please format your response strictly in this format without additional explanation:
<knowledge> [your knowledge of NetHack] </knowledge>
<analysis> [your one-sentence analysis of the message] </analysis>
<label> [FOO/BAR] </label>
\end{lstlisting}

\clearpage
\textbf{Prompt for \oni-ranking}

\begin{lstlisting}
I will present you with two short gameplay descriptions.
First, tell me about your knowledge of NetHack. Mention the goal of NetHack. Prefer agents that maximize the score in the game, for instance by killing monsters, collecting gold or going down the stairs in the dungeon.
Then, write an analysis describing the semantics of each description strictly using information from the descriptions (which may be empty) and your knowledge of NetHack.
Provide a comparative analysis based on first principles.
Finally, respond by explicitly declaring which one is the most likely to make some progress towards the goal, writing either ("best_description": 1), ("best_description": 2). You could also say ("best_description": None).

{{
"description_1":
"{}"
}}

{{
"description_2":
"{}"
}}
\end{lstlisting}

\subsection{Goal String and Prompts used in \Cref{sec:comp_cosine_bow}}

For the experiments in Section \ref{sec:comp_cosine_bow}, we use the goal strings from \cite{klissarovdoro2023motif} used to induce different behaviors in agents, such as the Gold Collector, Monster Slayer or Descender. These goal string explicitly aim to maximize one metric while minimize the others. The default goal string is the same used in \oni prompts. 
\label{app:goal_strings}

\subsubsection{Prompt used for \texttt{ELLM-BoW}}
\label{app:goal_string_cosine_bow}

\textbf{Default Goal} \begin{lstlisting}
Prefer agents that maximize the score in the game, for instance by killing monsters, collecting gold or going down the stairs in the dungeon.
\end{lstlisting}

\textbf{Combat Goal} \begin{lstlisting}
Prefer agents that engage in combat, for instance by killing monsters. But never prefer agents that collect ANY gold or that go down the dungeon.
\end{lstlisting}

\textbf{Gold Goal} \begin{lstlisting}
Prefer agents that maximize their gold. But never prefer agents that maximize the score in other ways (e.g., by engaging in combat or killing monsters) or that go down the dungeon.
\end{lstlisting}

\subsubsection{Prompt used for \oniretrieval}
\label{app:prompt_goal_string_oni}

The prompt used for \oni methods have been modified to reflect the change of goals, which we list in the next subsection. In addition to the changes of goal description, we use the same examples as in \Cref{app:prompts} but changed its label.

\textbf{Combat Goal}

\begin{lstlisting}
Context: I am playing NetHack and I will present you a message from the gameplay.

Instruction: First, tell me about your knowledge of NetHack. Mention the goal of NetHack.
Then, write an analysis describing the semantics of each message using your knowledge of NetHack. Finally, respond by assigning a score to the message, which measures whether the status represented by the message is helpful or unhelpful to make some progress towards the goal.
Helpful here does not mean whether it is informative to the player to make decisions or not, but whether it is helpful for the agent to solve the game.
I prefer agents engage in combat, for instance by killing monsters. But never prefer agents that collect ANY gold or that go down the dungeon.
For extremely helpful messages, assign a label FOO. For unhelpful, neutral, and lukewarm messages, assign a label BAR.

Example 1: "You kill the goblin!  Welcome to experience level 4" indicates that the player
has killed a goblin and gained experience. This led to a level up and increased abilities.
Gaining experience and leveling up are both extremely helpful and this message should be labelled as FOO.

Example 2: "You hear someone counting money." indicates that the player is close to a source of gold, as the sound of counting money suggests a nearby NPC or monster with treasure.
This is not helpful and should be labelled as BAR.

Example 3: "5 gold pieces." indicates you collected 5 gold coins. This is not aligned with our goal and should be labelled as BAR.

Example 4: "That door is closed." is neutral and should be labelled as BAR.

Please rate this message: {{"{}"}}

Constraint: Please format your response strictly in this format without additional explanation:
<knowledge> [your knowledge of NetHack] </knowledge>
<analysis> [your one-sentence analysis of the message] </analysis>
<label> [FOO/BAR] </label>
\end{lstlisting}

\textbf{Gold Goal}

\begin{lstlisting}[escapechar=@]
Context: I am playing NetHack and I will present you a message from the gameplay.

Instruction: First, tell me about your knowledge of NetHack. Mention the goal of NetHack.
Then, write an analysis describing the semantics of each message using your knowledge of NetHack. Finally, respond by assigning a score to the message, which measures whether the status represented by the message is helpful or unhelpful to make some progress towards the goal.
Helpful here does not mean whether it is informative to the player to make decisions or not, but whether it is helpful for the agent to solve the game.
I prefer agents that maximize their gold. But never prefer agents that maximize the score in other ways (e.g., by engaging in combat or killing monsters) or that go down the dungeon.
For extremely helpful messages, assign a label FOO. For unhelpful, neutral, and lukewarm messages, assign a label BAR.

Example 1: "You kill the goblin!  Welcome to experience level 4" indicates that the player
has killed a goblin and gained experience.  This is not aligned with our goal and should be labelled as BAR.

Example 2: "You hear someone counting money." indicates that the player is close to a source of gold, as the sound of counting money suggests a nearby NPC or monster with treasure. This is extremely helpful and should be labelled as FOO.

Example 3: "5 gold pieces." indicates you collected 5 gold coins. This is extremely helpful and should be labelled as FOO.

Example 4: "That door is closed." is neutral and should be labelled as BAR.

Please rate this message: {{"{}"}}

Constraint: Please format your response strictly in this format without additional explanation:
<knowledge> [your knowledge of NetHack] </knowledge>
<analysis> [your one-sentence analysis of the message] </analysis>
<label> [FOO/BAR] </label>
\end{lstlisting}

\clearpage
\section{Additional Results}
\label{sec:additional-results}

\subsection{LLaMA-3.1-70B-Instruct vs. LLaMA-3.1-8B-Instruct}
\label{app:llm_model_size}

\begin{figure}[H]
    \centering
    \includegraphics[width=0.75\linewidth]{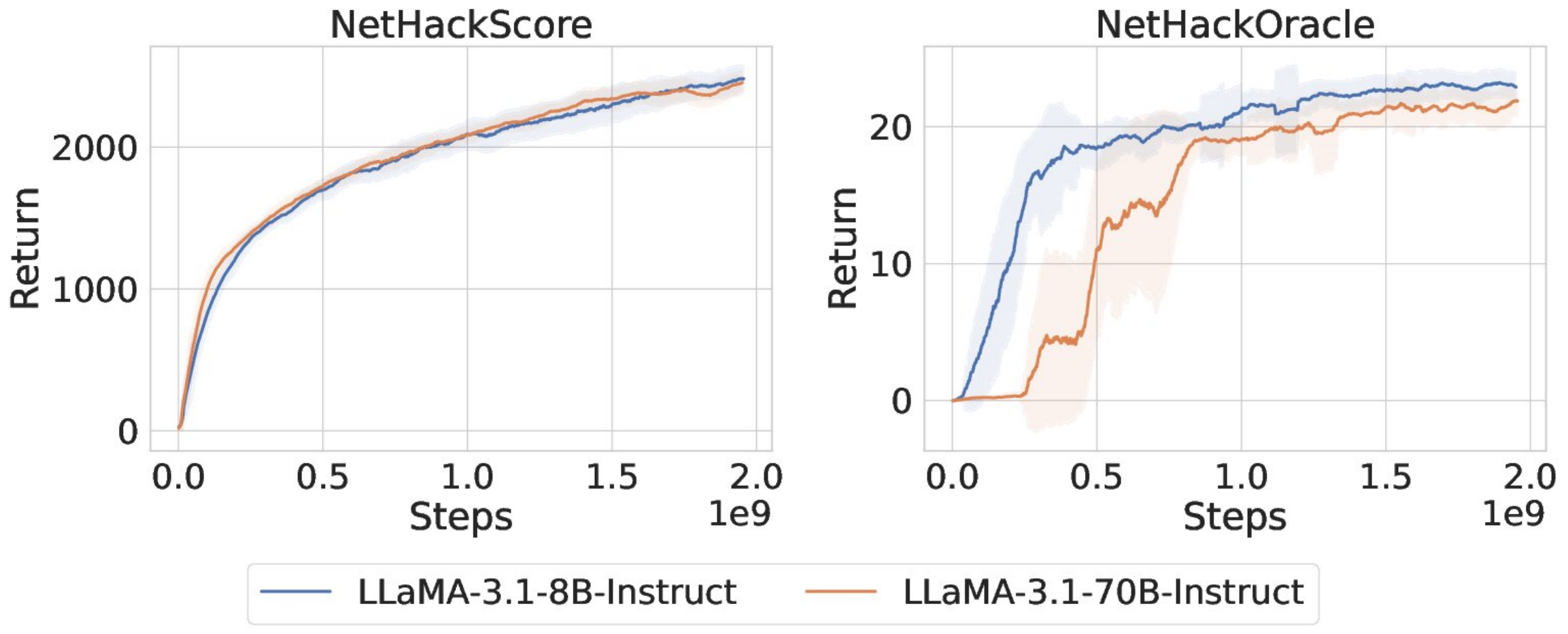}
    \caption{Performance of Motif using two different LLMs. Curves represent means and shaded regions represent standard errors over $5$ seeds. }
    \label{fig:motif_model_size}
\end{figure}

In \Cref{fig:motif_model_size}, we compare the performance of Motif using two different sized LLMs on \texttt{Score} and \texttt{Oracle} (LLaMA-3.1-8B-Instruct and LLaMA-3.1-70B-Instruct). Interestingly, we do not observe a significant difference between the two. This is in contrast to the previous work of \citep{klissarovdoro2023motif}, who found a significant difference between using LLaMA-2-70B-chat and LLaMA-2-7B-chat. This suggest that the smaller 8B model is sufficient for evaluating messages on these tasks, hence we use it in our experiments. 

\subsection{Effect of the Episodic Term for ELLM-BoW}
\label{app:episodic_bonus_cosine_bow}

\begin{figure}[H]
    \centering
    \includegraphics[width=0.85\linewidth]{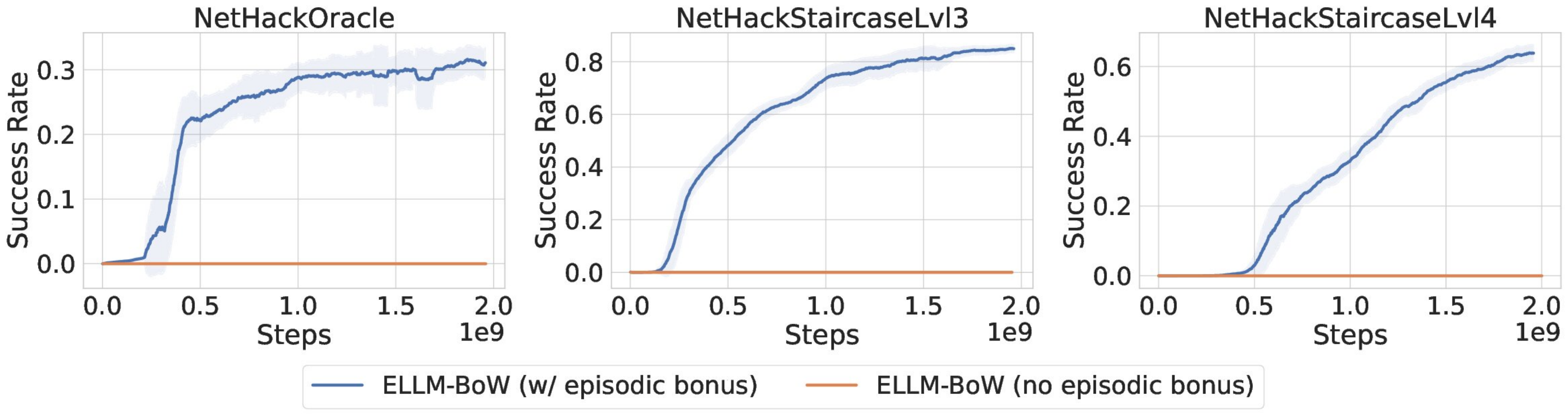}
    \caption{\texttt{ELLM-BoW} performs well when the intrinsic reward is normalized by an episodic-count based term as in~\eqref{eq:episodic_normalization}. Without it, the success rate is zero for all the three tasks.}
    \label{fig:cosine_bow_episodic}
\end{figure}

Our implementation of \texttt{ELLM-BoW} has included the episodic-count based normalization~\eqref{eq:episodic_normalization}, which is key to the performance of  \texttt{ELLM-BoW}.
\Cref{fig:cosine_bow_episodic} shows that directly using ELLM-BoW as in \citet{du2023guiding} \texttt{ELLM-BoW}, without the episodic term, failed to make progress in all three tasks.

\subsection{Additional Ablations}
\label{app:additional-ablations}

\label{app:a100_vs_v100}
\begin{figure}[H]
    \centering
    \includegraphics[width=0.55\linewidth]{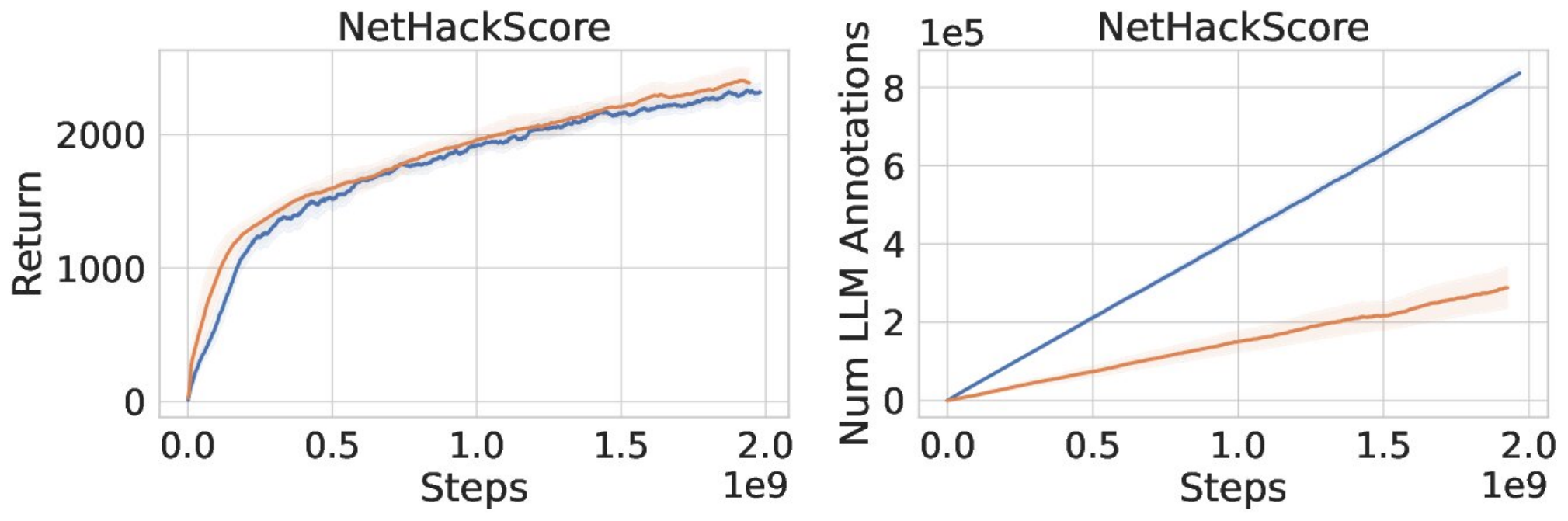}\\
    \includegraphics[width=0.55\linewidth]{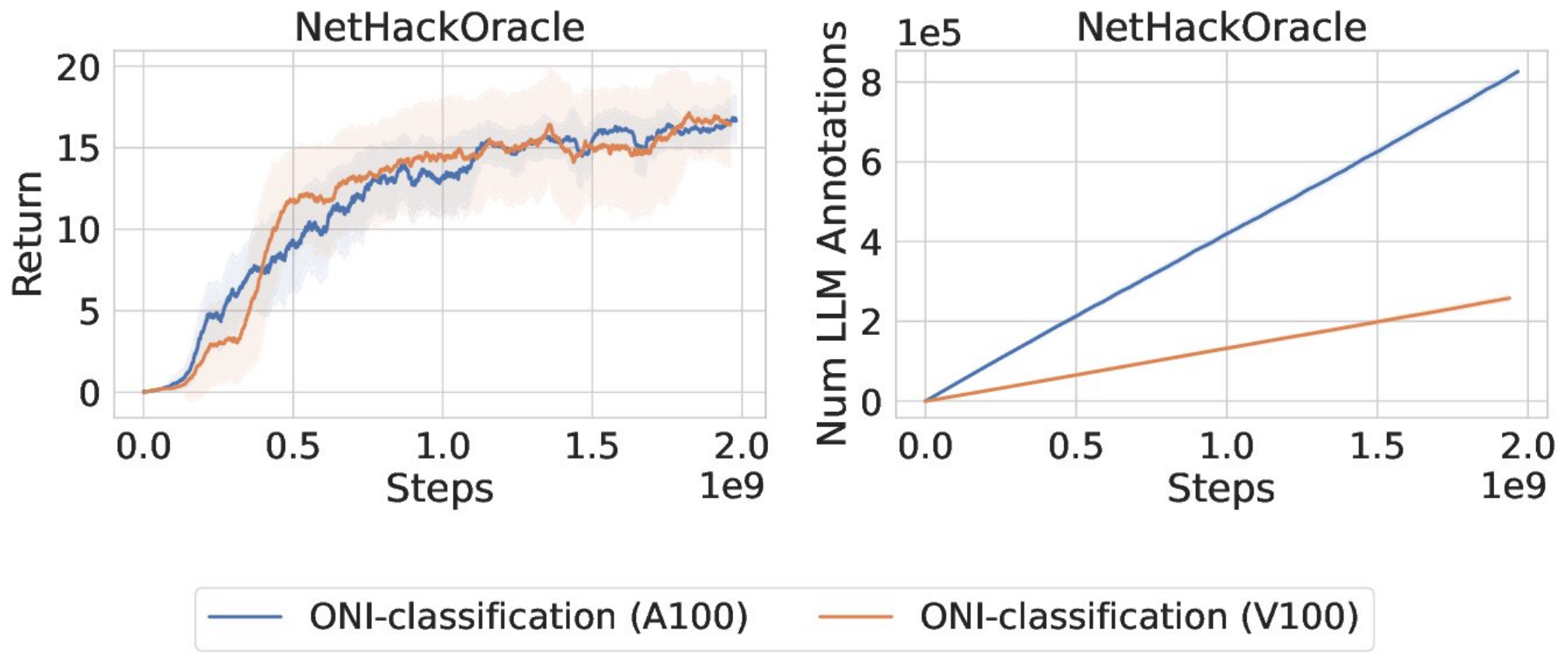}\\
    \caption{
        Performance of \onicls is comparable when the LLM server uses a Tesla A100-80GB or V100-32GB GPU.
    }
    \label{fig:a100_vs_v100}
\end{figure}

\begin{figure}[H]
    \centering
    \includegraphics[width=\linewidth]{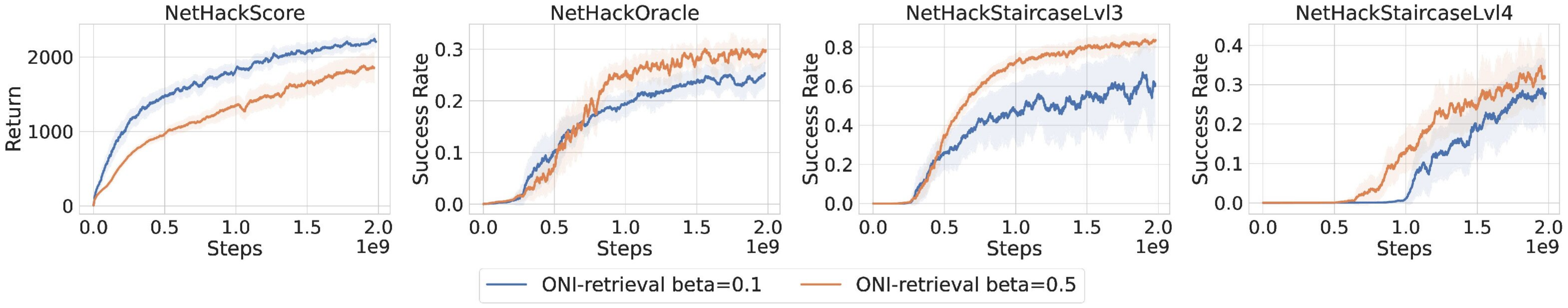}\\
    \includegraphics[width=\linewidth]{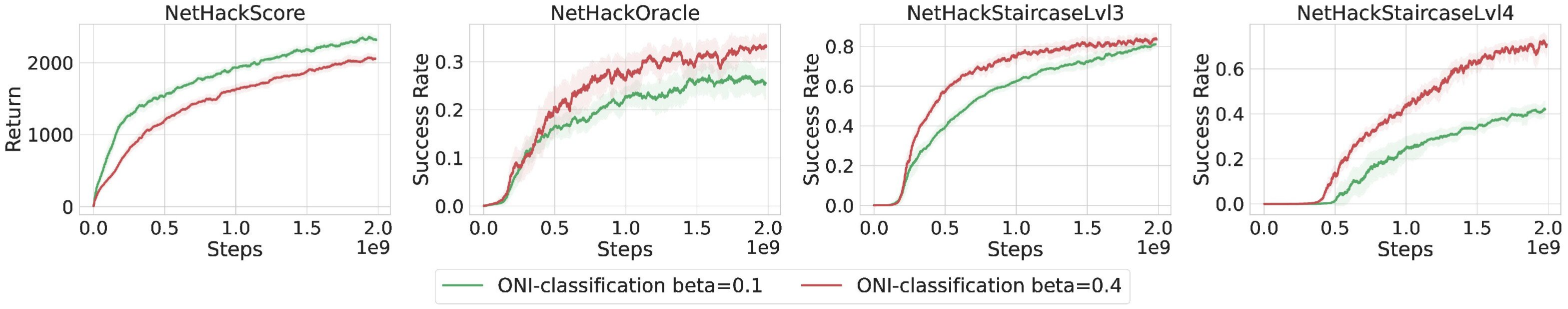}\\
    \includegraphics[width=\linewidth]{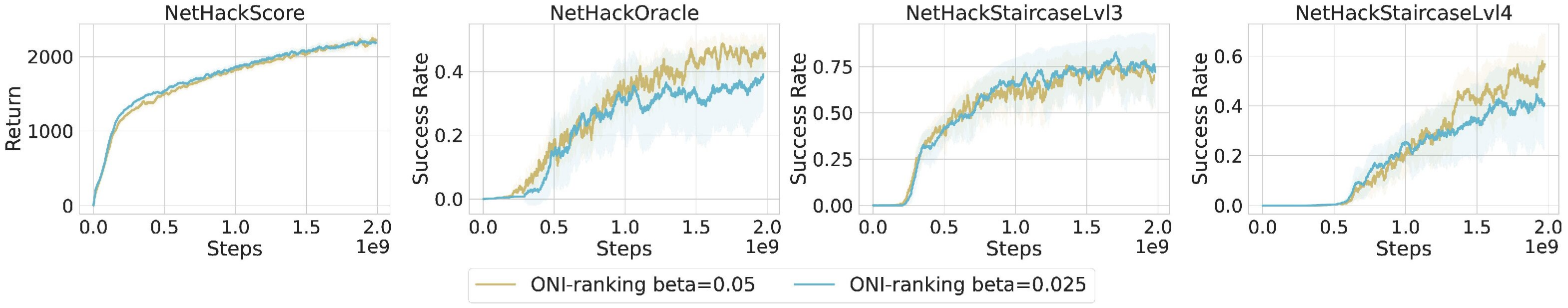}\\
    \caption{
        For \oniretrieval and \onicls, sparse reward tasks favor larger intrinsic reward coefficient $\beta$ while smaller values of $\beta$ lead to better results for
        the dense reward \texttt{Score} task. 
        For \oniranking, we do not observe much difference for the \texttt{Score} task, but the sparse reward tasks still slightly favors larger $\beta$. \
    }
    \label{fig:ablation_lrc}
\end{figure}

\begin{figure}[ht]
    \centering
        \includegraphics[width=0.9\linewidth]{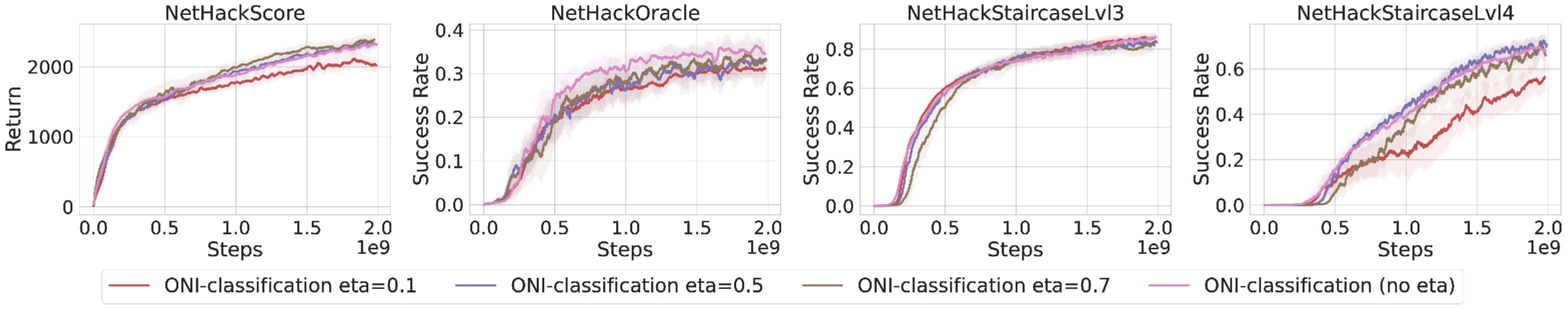}
        \includegraphics[width=0.9\linewidth]{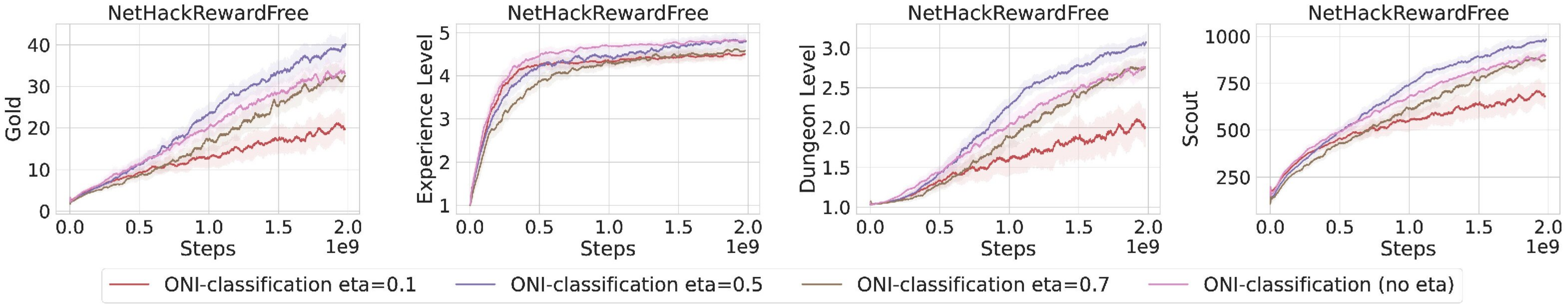}
    \caption{\textbf{(Top)} Performance of \onicls is robust to different choices of the classification threshold $\eta$ on the four NetHack tasks. 
    \textbf{(Bottom)} $\eta = 0.5$ marginally outperform the other values for training intrinsic-reward-only agents. The result when using $P(y_t=1|o_t)$ as reward \eqref{eq:oni_classification_no_eta} is 
    marked with legend \texttt{oni-classification (no eta)}.
    }
    \label{fig:ablation_eta}
\end{figure}

\textbf{Impact of Intrinsic Reward Coefficient $\beta$}
For \onicls and \oniretrieval, we have used different values of $\beta$ for the \texttt{Score} task and other sparse reward tasks in \Cref{sec:expr_main}.
Throughout our experiments, we have found that
for these two methods, 
larger values of $\beta$ lead to better performance for the sparse reward tasks, while smaller values of $\beta$ better balance between intrinsic and extrinsic rewards for the dense reward \texttt{Score} task. In comparison, \oniranking is relatively robust to this choice, where larger values of $\beta$ are still slightly favored for the sparse reward tasks.
\textbf{\onicls: the impact of the classification threshold} 
As described in \Cref{sec:methods}, \onicls predicts binary rewards by modeling $P(y_t=1|o_t)$ and then thresholding with $\eta$.
Instead of using binary reward, an alternative design choice is to use real valued reward
\begin{equation}
\label{eq:oni_classification_no_eta}
    \rint(o_t) = P(y_t=1|o_t),
\end{equation}
where $\rint(o_t) \in [0, 1]$ is the output of the reward classifier before thresholding. 
\Cref{fig:ablation_eta} shows that the performance of \onicls on the four NetHack tasks
is relatively robust to this hyperparameter $\eta$, and using $P(y_t=1|o_t)$ as the reward leads to similar performance.
As the training progresses, our reward model outputs values close to 0 or 1\footnote{To increase the diversity of the captions in the training dataset, we do not requery the LLM server if a caption is already annotated before. 
Therefore, every caption in our dataset only has a single label, resulting in this phenomenon.},
and we hypothesize this is the reason why the final performance remains comparable.
Moreover, the natural classification threshold $\eta=0.5$ %
marginally outperforms the other values in the reward-free setting. 

\paragraph{\oniranking: the impact of sampling strategy for LLM annotation and reward training}
Unlike \onicls that uses a LIFO queue to rank captions for LLM annotation, it is more subtle to design the most effective sampling strategy for \oniranking, where we need to construct pairs of captions to annotate and use for reward model training.
Here we study the effect of deduplicating captions before passing them to our pipeline. In either case, we maintain a message list $\mathcal{L}$, and sample pairs of captions $c_1, c_2 \sim \text{Uniform}(\mathcal{L} \times \mathcal{L})$ for annotation. The annotated message pairs are stored in another list, which is sampled from uniformly when training the reward model.
In the first option, each time the agent encounters a message, we check if it is already stored in $\mathcal{L}$ and only add it if not. This is similar to the approach used in \onicls. 
In the second option, we simply add all captions encountered by the agent into $\mathcal{L}$, regardless of whether they have already been seen before. This approach is similar to that taken by the original Motif work, which does not perform any deduplication of the offline dataset. 

\begin{figure}[ht]
    \centering
    \includegraphics[width=\linewidth]{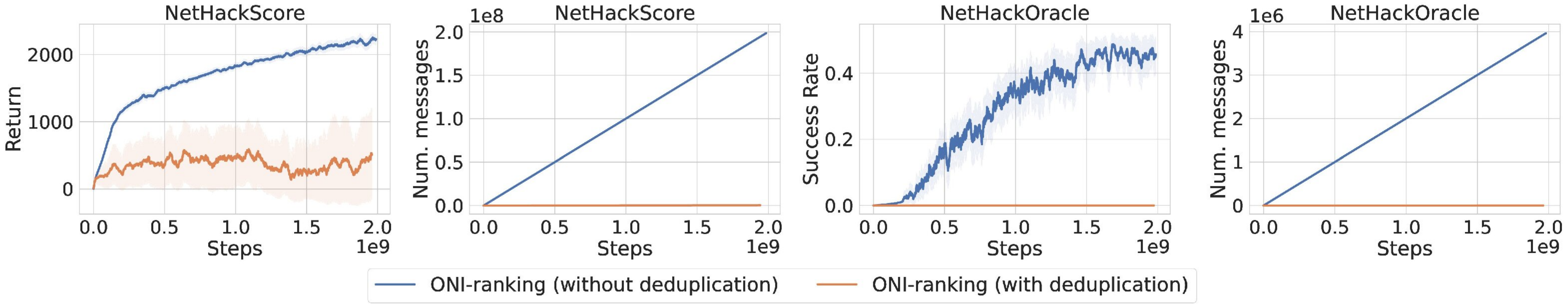}
    \caption{ 
    Performance and number of captions in the replay buffer for agents trained with and without message deduplication. Keeping the natural distribution of captions, and not deduplicating, is important to enable our ranking-based method to learn. 
    }
    \label{fig:ablation_ranking_offset_vs_thresholding}
\end{figure}

\Cref{fig:ablation_ranking_offset_vs_thresholding} shows the performance and the number of captions stored in the replay buffer for both schemes. We see that for the deduplicated variant, the number of captions does not grow past a certain point (approximately $70k$ unique captions). For the non-deduplicated variant, the number of captions keeps growing linearly over time. The deduplicated variant fails to learn, which highlights the important effect which the annotation dataset can have. We hypothesize that the deduplicated variant may undersample captions that occur frequently in the agent's experience, for which it is important to reliably estimate reward. For example, the blank message occurs very frequently during policy learning, but is only included in a small fraction of the pairs sent to the LLM for annotation, since it is sampled with the same probability as the other $\sim 70k$ other captions. In contrast, the duplicated variant samples the blank message and other frequent captions with much higher probability. Still, this remains a simple strategy, and designing more sophisticated sampling mechanisms (for example, that account for the epistemic uncertainty of the reward model) would be an interesting direction for future work.

\section{Additional Related Work}
\label{app:additional_related_work}

\paragraph{LLM for RL Broadly}
Another way of leveraging the prior knowledge encoded in LLMs for decision making is to use the LLM directly as a policy. This approach has been successfully used in robotics \citep{ahn2022icanisay, driess2023palme},
as well as open-ended exploration in MineCraft \citep{wang2024voyager}. Both settings require the LLM to operate at a higher level of abstraction, by having it call upon a set of semantically grounded skills which handle the low-level sensorimotor activity. These are in turn produced by imitation learning on expert trajectories or hardcoded APIs.
\citet{jeurissen2024playing} prompt the LLM to choose a predefined skill to play NetHack.
The prompts are constructed to represent past events, current observation, and the task description and hardcoded available skills are also included.
More references on LLMs for decision making can be found in the survey paper of \citet{cao2024survey}.

\end{document}